\documentclass{article}

\usepackage{microtype}
\usepackage{graphicx}
\usepackage{subcaption}
\usepackage{booktabs}

\usepackage{hyperref}

\usepackage[preprint]{icml2026}

\usepackage{amsmath}
\usepackage{amssymb}
\usepackage{mathtools}
\usepackage{amsthm}

\usepackage{multirow}
\usepackage{CJKutf8}
\usepackage{tcolorbox}

\usepackage[capitalize,noabbrev]{cleveref}

\theoremstyle{plain}

\theoremstyle{definition}

\theoremstyle{remark}

\usepackage[textsize=tiny]{todonotes}

\usepackage{xcolor}
\usepackage[table]{xcolor}
\definecolor{lightpurple}{RGB}{230,230,255}

\icmltitlerunning{From Myopic Selection to Long-Horizon Awareness: Sequential LLM Routing for Multi-Turn Dialogue}

\begin{document}

\twocolumn[
  \icmltitle{From Myopic Selection to Long-Horizon Awareness:\\ Sequential LLM Routing for Multi-Turn Dialogue}
  \icmlsetsymbol{equal}{*}

  \begin{icmlauthorlist}
    \icmlauthor{Jiarui Zhang}{1}
    \icmlauthor{Xiangyu Liu}{2}
    \icmlauthor{Yong Hu}{2}
    \icmlauthor{Chaoyue Niu}{1}
    \icmlauthor{Hang Zeng}{1}
    \icmlauthor{Shaojie Tang}{3}
    \icmlauthor{Fan Wu}{1}
    \icmlauthor{Guihai Chen}{1}
  \end{icmlauthorlist}

  \icmlaffiliation{1}{School of Computer Science, Shanghai Jiao Tong University, Shanghai, China
}
  \icmlaffiliation{2}{WeChat, Tencent Inc, Beijing, China}
  \icmlaffiliation{3}{University at Buffalo, New York, United States}

  \icmlcorrespondingauthor{Chaoyue Niu}{rvince@sjtu.edu.cn}

  \vskip 0.3in
]

\printAffiliationsAndNotice{}

\begin{abstract}
  Multi-turn dialogue is the predominant form of interaction with large language models (LLMs). While LLM routing is effective in single-turn settings, existing methods fail to maximize cumulative performance in multi-turn dialogue due to interaction dynamics and delayed rewards. To address this challenge, we move from myopic, single-turn selection to long-horizon sequential routing for multi-turn dialogue. Accordingly, we propose DialRouter, which first performs MCTS to explore dialogue branches induced by different LLM selections and collect trajectories with high cumulative rewards. DialRouter then learns a lightweight routing policy from search-derived data, augmented with retrieval-based future state approximation, enabling multi-turn routing without online search. Experiments on both open-domain and domain-specific dialogue tasks across diverse candidate sets of both open-source and closed-source LLMs demonstrate that DialRouter significantly outperforms single LLMs and existing routing baselines in task success rate, while achieving a superior performance-cost trade-off when combined with a cost-aware reward.
\end{abstract}

\section{Introduction}
\begin{figure}[t]
    \centering

    \begin{subfigure}{\linewidth}
        \centering
        \includegraphics[width=1.0\linewidth]{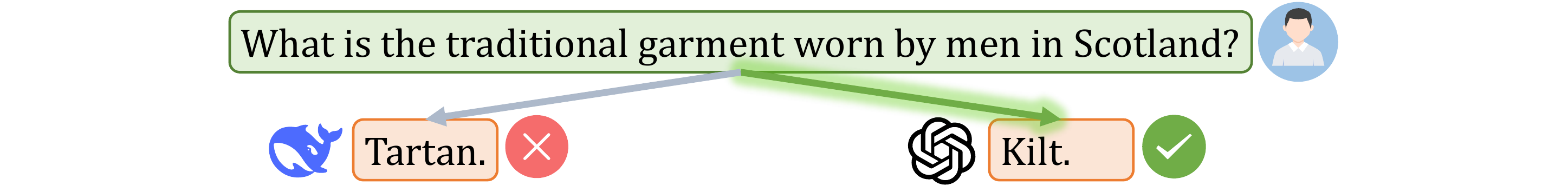}
        \caption{LLM Routing in Single-turn Interaction.}
        \label{fig:top}
    \end{subfigure}

    \hfill

    \begin{subfigure}{\linewidth}
        \centering
        \includegraphics[width=1.0\linewidth]{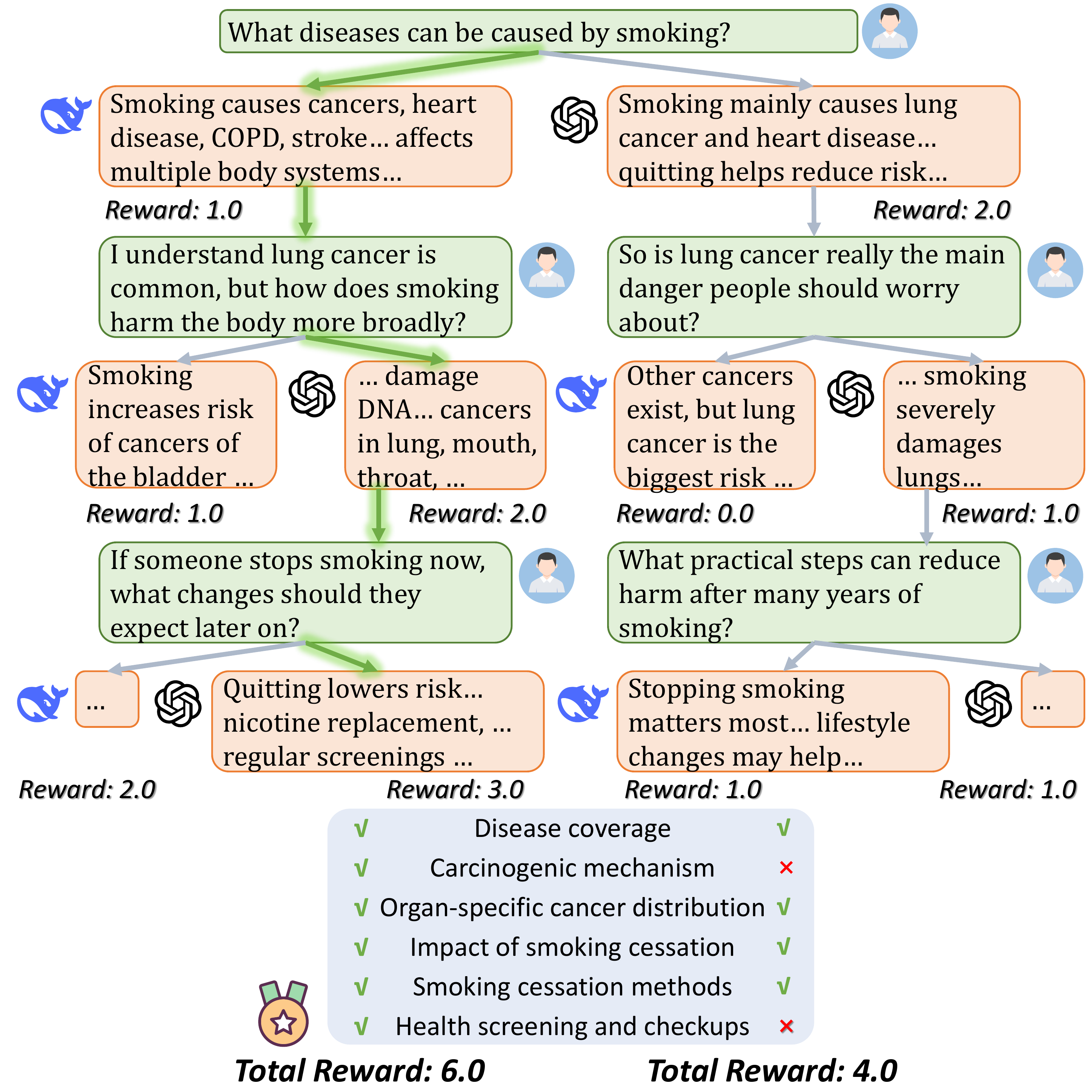}
        \caption{LLM Routing in Multi-turn Dialogue.}
        \label{fig:bottom}
    \end{subfigure}

    \caption{Illustration of LLM routing in single-turn interaction (a) and multi-turn dialogue (b). Green arrows indicate ideal routing decisions. In multi-turn dialogue, different LLM selections induce different user queries. Rewards are distributed across turns, while the total reward reflects overall user intent fulfillment.}
    \label{fig:intro}
    \vskip -0.2in
\end{figure}

Multi-turn dialogue is the dominant paradigm of user interaction with large language models (LLMs). In everyday general-purpose applications such as ChatGPT \cite{achiam2023gpt}, Gemini \cite{comanici2025gemini}, Qwen \cite{yang2025qwen3} and DeepSeek \cite{liu2024deepseek}, user engagement rarely consists of isolated queries; instead, users naturally participate in iterative conversations, providing clarification, feedback, and refinements as tasks progress \cite{yi2025survey}. This multi-turn interaction pattern is also evident across diverse LLM-driven vertical applications, including e-commerce customer service \cite{li2024ecomgpt}, medical consultation \cite{zhang2023huatuogpt}, educational tutoring \cite{liu2024socraticlm}, emotional support \cite{chen2023soulchat}, and programming assistance \cite{nam2024using}.

Multi-turn dialogues encompass diverse user intents that impose varying demands on LLM capabilities. Meanwhile, different LLMs exhibit complementary strengths in instruction following, knowledge coverage, and reasoning \cite{liang2022holistic}. To leverage this complementarity, LLM routing selects the most suitable LLM for each query from a pool of pre-deployed LLMs \cite{chen2025harnessing} and has been shown to outperform strong individual LLMs \cite{hu2024routerbench, chen2024routerdc, zhang2025avengers}. However, existing routing methods are developed and evaluated in single-turn interaction settings, rather than multi-turn dialogue. This gap raises a natural question: \textit{How should LLM routing be designed for multi-turn dialogue, where model selection decisions unfold sequentially in the evolving context?}

As illustrated in Figure~\ref{fig:top}, in single-turn interaction, users seek a correct answer to a clearly defined query, and the router simply selects the best-performing LLM. In contrast, in multi-turn dialogue, users progressively advance a set of interrelated core intents across turns, with success defined by whether these intents are collectively fulfilled throughout the dialogue. As shown in Figure~\ref{fig:bottom}, multi-turn dialogue routing is an evolving decision process, where the router selects a LLM at each turn, receiving immediate rewards for intent progression and a total reward reflecting overall intent fulfillment. Notably, selecting different LLMs under the same context can elicit different follow-up user queries in subsequent turns, causing the dialogue to evolve along distinct trajectories and highlighting its \textbf{dynamic} nature. Moreover, overall intent fulfillment can only be assessed after the dialogue unfolds, exhibiting clear \textbf{delayed rewards}.

Due to the dynamic and delayed-reward nature of multi-turn dialogue, per-turn observable signals often fail to capture the long-term impact of LLM selection. As illustrated in Figure~\ref{fig:bottom}, a locally optimal choice at a given turn does not necessarily yield a globally optimal dialogue outcome. Beyond performance, inference costs accumulate over turns, and frequent model switching disrupts cache reuse\footnote{\url{https://platform.openai.com/docs/guides/prompt-caching}}, making locally cost-optimal routing suboptimal at the dialogue level. We thus argue for a shift from myopic, per-turn selection to long-horizon LLM routing in multi-turn dialogue.

In this work, we propose \textbf{DialRouter}, a sequential LLM routing framework for multi-turn dialogue. We first employ Monte Carlo Tree Search (MCTS) together with a user simulator to explicitly explore dialogue branches induced by different LLM selections. By rolling out complete dialogue trajectories and evaluating their cumulative returns with a reward model, we collect routing trajectories with high long-term returns. The reward model adopts a checklist-based formulation, well suited to fine-grained user intent fulfillment in multi-turn dialogue. Offline policy learning is then performed on these trajectories via behavior cloning. Since relying solely on the current dialogue state is insufficient to infer the long-term impact of routing decisions, motivated by the idea that approximated future states can provide guidance on potential dialogue evolution and ease routing decisions, we introduce a retrieval-based future state approximation derived from search trajectories to assist routing. Through policy learning, DialRouter distills long-horizon knowledge from search into an efficient routing policy, enabling high-quality multi-turn routing without online search.

Our main contributions are summarized as follows:
\begin{itemize}
    \item To the best of our knowledge, we are the first to advance LLM routing from myopic single-turn selection to long-horizon routing for multi-turn dialogue.
    \item We propose \textbf{DialRouter}, which leverages MCTS-discovered trajectories and behavior cloning with retrieval-based future state approximation to enable search-free multi-turn routing.
    \item We extend the reward formulation to incorporate LLM inference costs and KV cache reuse, facilitating performance-cost trade-offs.
    \item Extensive experiments across three dialogue domains and multiple open- and closed-source LLM candidate sets show that DialRouter consistently outperforms the strongest single LLMs and existing routing baselines in task success rate, while achieving superior performance--cost trade-offs under cost-aware settings.
\end{itemize}

\section{Related Work}
\textbf{LLM Routing.} Existing LLM routing methods mainly target single-turn interaction. Representative works such as RouteLLM \cite{ong2025routellm} and EmbedLLM \cite{zhuang2025embedllm} learn compact query and model representations via matrix factorization to predict routing correctness. RouterDC \cite{chen2024routerdc} applies dual contrastive learning to capture query–model compatibility. RouterBench \cite{hu2024routerbench} systematically evaluates a range of simple yet effective routing strategies, including KNN-based methods and lightweight MLPs. Avengers \cite{zhang2025avengers} proposes a training-free routing approach that performs model selection through query embedding clustering and response score ranking. Router-R1 \cite{zhang2025router} decomposes complex queries into sub-queries and performs routing sequentially over these components. GMTRouter \cite{xie2025gmtrouter} leverages user-LLM interaction histories to enable personalized routing decisions. Although Blending \cite{lu2024blending} considers multi-turn dialogue with multiple LLMs, it adopts random response selection instead of explicitly learning routing decisions. In contrast, DialRouter adopts a holistic multi-turn dialogue perspective and formulates LLM routing as a long-horizon decision-making problem, explicitly modeling the impact of LLM selection on future dialogue evolution and final user intent fulfillment.

\textbf{Multi-turn Dialogue Planning.} Different from LLM routing, which focuses on selecting among multiple LLMs, existing work on multi-turn dialogue planning typically based on a single LLM and applies search or learning methods over a limited space of prompting strategies to accomplish predefined cooperative or adversarial dialogue tasks. GDP-Zero \cite{yu2023prompt} combines MCTS with LLM prompting to optimize policy selection. PPDPP \cite{deng2024plug} equips LLMs with an external planner and trains the planner via RL for strategy prediction. Inspired by AlphaZero \cite{silver2017mastering}, methods such as DPDP \cite{he2024planning} and DMNA \cite{liu2025dual} integrate MCTS with learning algorithms to enhance policy networks for dialogue planning. Although effective in dialogue control, these methods focus on selecting among predefined prompting strategies, rather than choosing which LLM should be used. In more general dialogue settings, heterogeneous LLM capabilities elevate planning to the level of LLM selection. In this work, we study the LLM routing problem in multi-turn dialogue.

\section{Preliminaries}
\subsection{Multi-Turn Dialogue and User Intent Fulfillment}
A multi-turn dialogue can be represented as an interaction sequence $\tau_T = [x_1, y_1, x_2, y_2, \dots, x_T, y_T]$, where $x_t$ and $y_t$ denote the user input and the system response at turn $t$, respectively, and $T$ is the terminal turn of the dialogue.

To characterize the dialogue-level fulfillment of user intents in $\tau_T$, inspired by fine-grained evaluation criteria \cite{cook2024ticking, viswanathan2025checklists}, we explicitly decompose the user’s high-level goal into a set of assessable task requirements, forming a dialogue-level checklist $C = [c_1, c_2, \dots, c_n]$, where each item $c_i$ corresponds to a key constraint or sub-goal associated with the user’s intent over the course of the dialogue. At the end of the dialogue, each checklist item is evaluated based on the overall dialogue trajectory as \emph{fulfilled}, \emph{partially fulfilled}, or \emph{unfulfilled}, and is assigned a score $r_i^{(T)} \in \{1, 0.5, 0\}$, respectively, yielding the checklist completion score $R_T = \sum_{i=1}^{n} r_i^{(T)}$. The \emph{partially fulfilled} signal captures incremental intent progress that would be missed by a binary reward. Checklist construction details are provided in Appendix~\ref{appena3}.

\subsection{Multi-Turn Routing Problem Formulation}
LLM routing in multi-turn dialogues is formulated as a sequential decision-making problem, where at each turn the router selects an LLM from a candidate set $\mathcal{M} = \{M_1, \dots, M_N\}$ to generate a system response with the objective of maximizing the dialogue-level completion score $R_T$. At turn $t$, the router observes the current state $s_t = [\tau_{t-1}, x_t]$, consisting of the dialogue history up to the previous turn $\tau_{t-1}$ and the current user input $x_t$. Given the state $s_t$, the router selects a LLM from $\mathcal{M}$ to produce a system response. We denote this selection as the action $a_t$, and the response is generated as $y_t = M_{a_t}(s_t)$, after which the dialogue history is updated to $\tau_t = [\tau_{t-1}, x_t, y_t]$. The user subsequently issues the next query $x_{t+1}$ based on the current dialogue $\tau_t$ and the router observes the updated state $s_{t+1} = [\tau_t, x_{t+1}]$, entering the next turn of LLM selection.

 Since checklist items may be progressively addressed over the dialogue, we define $R_t$ as the checklist completion score at $\tau_t$ (with $R_0 = 0$), and define the immediate reward for action $a_t$ in state $s_t$ as the incremental checklist completion,
\begin{equation}\label{rewardf}
    \mathcal{R}(s_t, a_t) = R_t - R_{t-1}.
\end{equation}
Under this formulation, the objective of multi-turn LLM routing is to learn a routing policy $\pi_R$ that maximizes the expected cumulative reward,
\begin{equation}
    \max_{\pi_R} \; \mathbb{E}\!\left[\sum_{t=1}^{T} \mathcal{R}(s_t, a_t)\right].
\end{equation}
This objective is equivalent to maximizing the dialogue-level completion score $R_T$, since $R_T$ admits an additive decomposition across turns as $R_T = \sum_{t=1}^{T} (R_t - R_{t-1})$.

In contrast, single-turn routing focuses on selecting, at each interaction, the LLM that maximizes the immediate reward, without accounting for the impact of LLM choices on future interactions. In multi-turn dialogues, however, LLM response $y_t$ influences subsequent user behavior $x_{t+1}$, rendering routing strategies based only on immediate rewards inherently myopic and often suboptimal with respect to dialogue-level objectives. Only in the special case where future user inputs are independent of the preceding dialogue context does multi-turn dialogue LLM routing degenerate into a set of independent single-turn routing problems.

\section{DialRouter}
\begin{figure}[t]
    \centering
    \includegraphics[width=0.95\linewidth]{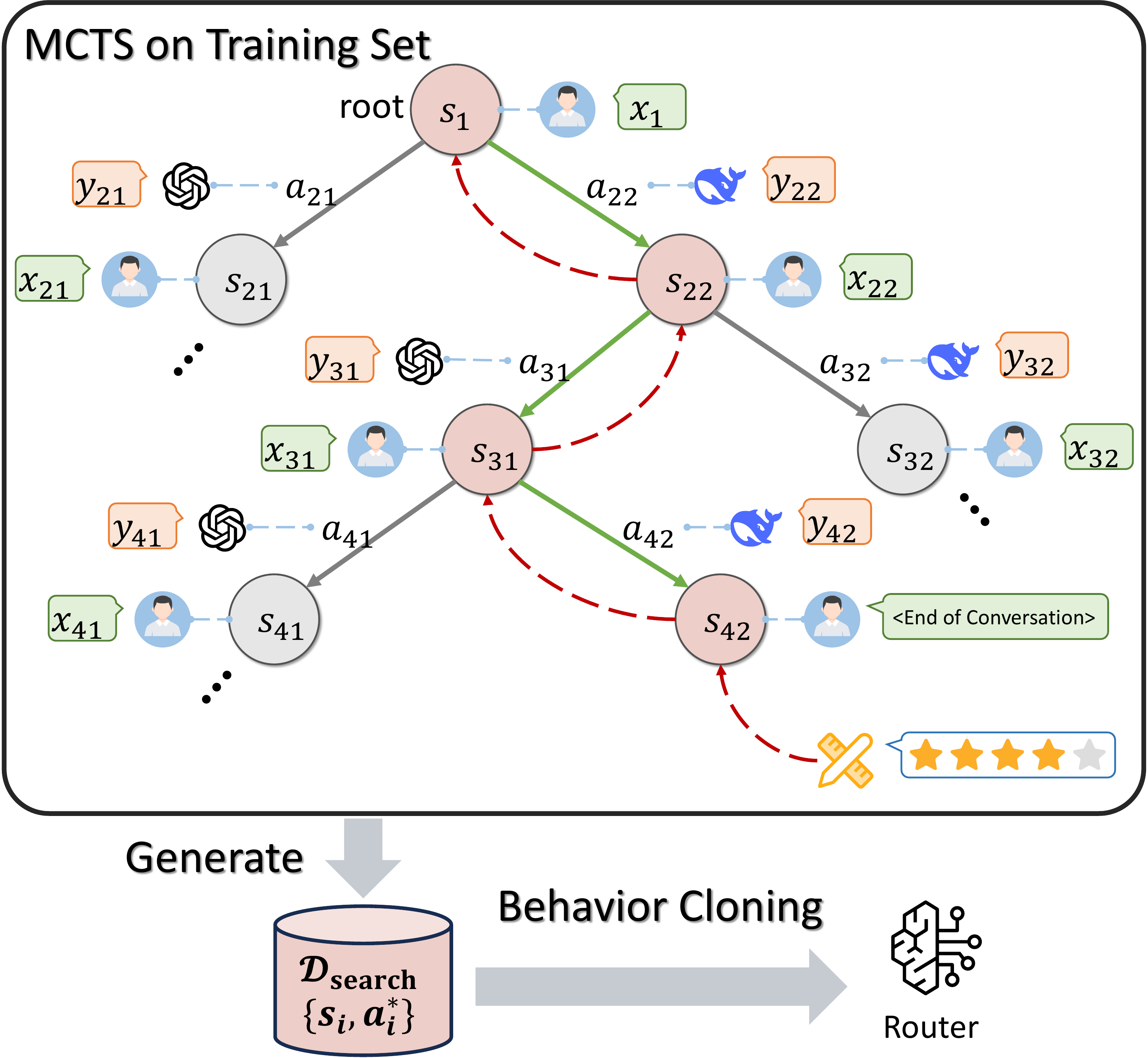}
    \caption{Overview of DialRouter. MCTS with a user simulator and reward model is used to explore dialogue branches and collect high-quality routing trajectories, which are then used for offline policy learning via behavior cloning to train a search-free router.}
    \label{fig:dial}
    \vskip -0.1in
\end{figure}

\begin{figure}[t]
    \centering
    \includegraphics[width=1.0\linewidth]{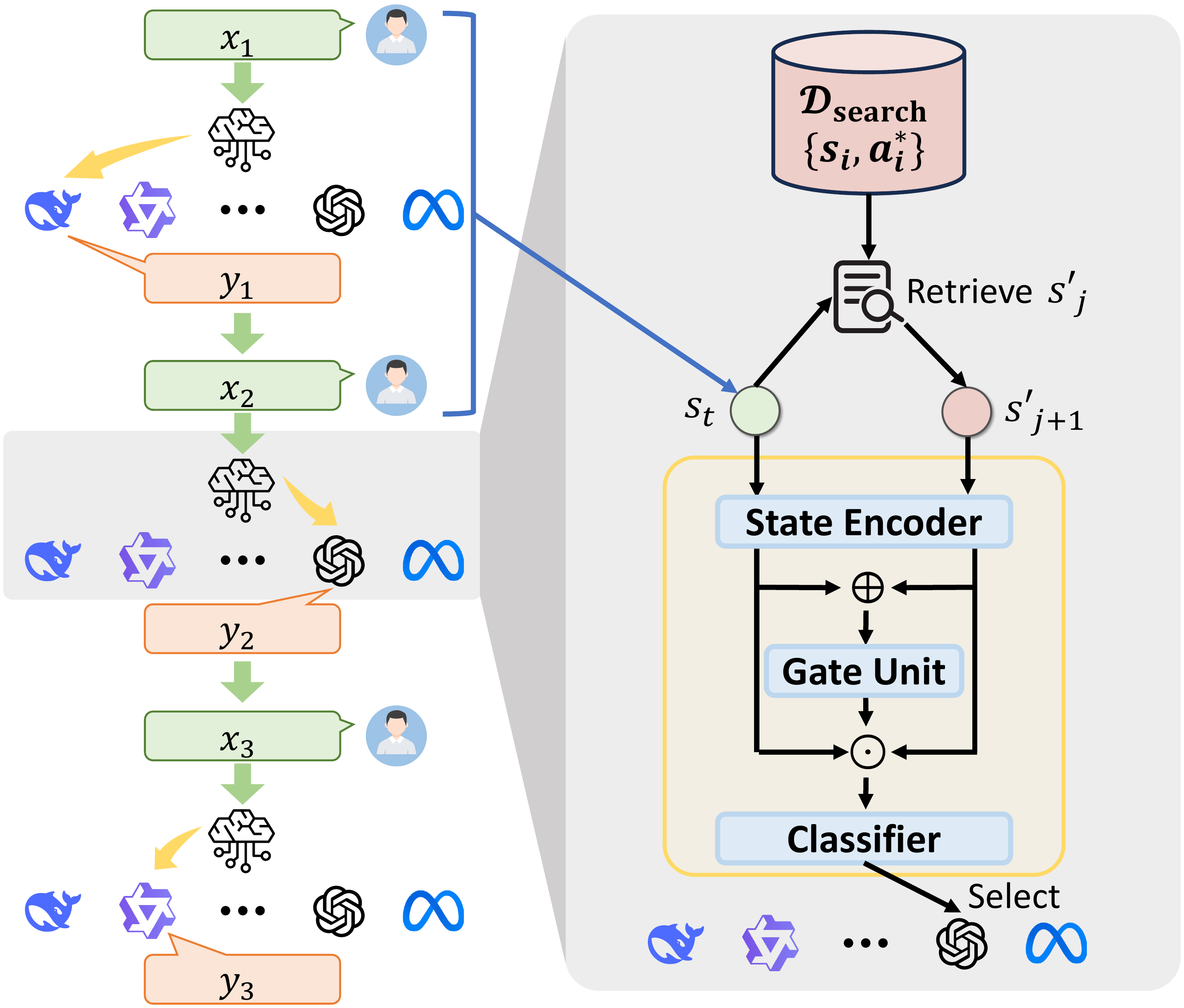}
    \caption{Routing process of DialRouter in multi-turn dialogue. At each turn, the current dialogue state $s_t$ and a retrieved future state approximation $s'_{j+1}$ from $\mathcal{D}_{\text{search}}$ are jointly encoded and fused to predict the LLM selection.}
    \label{fig:dial1}
\end{figure}

In this section, we present \textbf{DialRouter}, an LLM routing framework for multi-turn dialogue, as illustrated in Figure~\ref{fig:dial} and Figure \ref{fig:dial1}. Section~\ref{sec41} describes search-derived trajectory generation, where MCTS with a user simulator is used to explore dialogue branches induced by different LLM selections and collect routing decision trajectories with high cumulative rewards. Section~\ref{sec42} describes routing policy learning, where a routing model is trained via behavior cloning on the expert data and augmented with a retrieval-based future state approximation, enabling multi-turn routing without search at inference time.
\subsection{Simulation-Based Search Trajectory Generation}\label{sec41}

\textbf{User Query Simulation.} In multi-turn dialogue settings, routing decisions exhibit strong dynamics, as responses generated by different LLMs at the same turn can induce different user behaviors in subsequent turns, leading the dialogue to evolve along different trajectories and affecting the overall dialogue outcome. To capture this dynamic interaction process, we adopt an LLM-based user simulator to generate subsequent user responses. Specifically, the user simulator is responsible for producing the next user query and determining dialogue termination; conditioned on a predefined user profile $p_{\text{usr}}$ and the current dialogue history $\tau_t$, it generates the next user input as $x_{t+1} = M_{\text{usr}}(p_{\text{usr}}, \tau_t)$.

\textbf{LLM Response Branch Exploration.} Given a mechanism for simulating subsequent user queries, the remaining challenge is to systematically explore the dialogue branches induced by different LLM responses and assess their long-term impact on dialogue outcomes. We adopt MCTS to search over routing decision sequences. In the search process, each dialogue state $s_t$ is treated as a node in the search tree, with the current state $s_{\text{root}}$ as the root. For a given state, candidate LLMs correspond to different action branches, which are expanded by invoking the selected model to generate a system response. Each generated response is evaluated by an LLM-based reward model: given the task checklist $C$, the current dialogue state $s_t$, and the system response $y_t$, the reward model outputs per-item completion scores $[r_1^{(t)}, \dots, r_n^{(t)}] = M_{\text{rew}}(C, s_t, y_t)$, which form the basis for computing path-level returns. A user simulator then generates the next user input, driving the dialogue state to the next turn. By repeatedly expanding branches in the simulated environment, evaluating turn-level completions with the reward model, and advancing dialogue evolution via the user simulator, MCTS compares the cumulative returns of dialogue paths and identifies routing decision paths with superior overall performance. For each dialogue task in the training set, MCTS yields a routing decision sequence, which is decomposed into state--action pairs to construct a search-derived dataset $\mathcal{D}_{\text{search}}=\{(s_t, a_t^*)\}$, where $a_t^*$ denotes the routing action selected by MCTS at state $s_t$. 

Details of the user simulator, reward model, and MCTS are provided in Appendices~\ref{appenb} and~\ref{append}.

\subsection{Search-Guided Routing Policy Learning}\label{sec42}
\textbf{Retrieval-based Future State Approximation.} The routing model $\pi_R(a|s)$ takes the current dialogue state as input and outputs a predicted probability distribution over actions. However, in the search-derived data, the optimal action $a^*$ is obtained through search and corresponds to the dialogue branch that maximizes future cumulative rewards, making it challenging for the routing model to directly learn such long-horizon dependencies. Inspired by prior work \cite{pascanu2017learning, hafner2019dream}, we posit that if the router could ``imagine'' an ideal future dialogue state when making the current decision, even in the form of a coarse-grained semantic representation, the difficulty of routing would be substantially reduced. To this end, we introduce a long-term-aware state modeling strategy that augments the current state $s_t$ with an approximate future state $\tilde{s}_{t+1}$, capturing the potential trend of future dialogue evolution.

However, the real next state $s_{t+1}$ is unobservable, and predicting future states via parameterized generation in natural language or semantic space is difficult for a lightweight routing model. Therefore, we adopt a retrieval-based approximation. For a given current state $s_t$, we employ a fixed retriever $\mathrm{Ret}$ to retrieve a semantically similar state
\begin{equation}
    s_j' = \mathrm{Ret}(\mathcal{D}_{\text{search}}, s_t)
\end{equation}
from the search-derived dataset $\mathcal{D}_{\text{search}}$, and use its successor state $s'_{j+1}$ along the corresponding search-derived trajectory as an approximation of the future evolution of $s_t$. This approximate future state is incorporated as an auxiliary signal during training and is likewise used at inference time, enabling the routing policy to capture potential future dialogue trends under limited current semantic information.

\textbf{Routing Model Architecture.} As shown in Figure~\ref{fig:dial1}, We use a dialogue state encoder $E(\cdot)$ to encode the current state and the retrieved approximate future state as
\begin{equation}
    e_t = E(s_t), \quad \tilde{e}_{t+1} = E(s'_{j+1}).
\end{equation}
To balance the information from the current state and the potentially noisy retrieved future state, we introduce a trainable gating mechanism for adaptive fusion:
\begin{equation}
    g = \sigma\bigl(W_G [e_t, \tilde{e}_{t+1}]\bigr), \quad
h = g \odot e_t + (1-g) \odot \tilde{e}_{t+1}.
\end{equation}
The fused representation $h$ is then fed into the classification head $\mathrm{Cls}(\cdot)$ to produce the final routing policy output $\pi_R(\cdot|s_t)$, based on which the candidate model is selected.

\textbf{Long-term-aware Supervision.} As the training dataset $\mathcal{D}_\text{search}$ is obtained via planning with respect to long-term cumulative rewards, it provides supervision signals that are aligned with long-term outcomes, enabling long-term-aware learning of the routing policy. We adopt a behavior cloning approach, treating the MCTS routing policy $\pi_{\text{MCTS}}$ as the expert and training the routing model $\pi_R$ to imitate its decision behavior. The training objective is to minimize the cross-entropy loss between $\pi_R$ and the $\pi_\text{MCTS}$ actions:
\begin{equation}
    \min_\theta -\frac{1}{|\mathcal{D}_{\text{search}}|}\sum_{(s,a)\in\mathcal{D}_{\text{search}}} \log \pi_R(a \mid s;\theta).
\end{equation}
In this way, the router can absorb the long-horizon planning knowledge implicit in MCTS during training, and perform long-term decision at inference time without explicit search.

\section{Extending Reward Function with Cost}
LLM routing decisions jointly impact dialogue quality and monetary invocation cost, motivating the incorporation of cost modeling into the reward function. To this end, we extend the original reward design with an explicit cost component. Rather than treating cost as a simple linear function, our cost formulation is directly grounded in the publicly available API pricing rules\footnote{\url{https://openai.com/index/api-prompt-caching}} of major cloud providers.

Specifically, closed-source LLMs typically employ a KV cache mechanism to store historical context: when the same model is invoked in consecutive turns, cached context tokens are charged at a lower cost, whereas switching models invalidates the cache\footnote{Due to the complexity of real-world dialogues, we consider the worst-case scenario for model switching, in which the KV cache is assumed to be completely invalidated.}, requiring the entire dialogue history to be re-encoded and billed at the standard input cost. As a result, the invocation cost in multi-turn dialogue routing depends not only on which model is selected, but also on the sequence of model choices.

For each candidate model $M_i$, the response cost consists of two components: output and input cost. Let $c_\text{out}^{(i)}, c_\text{in}^{(i)}, c_\text{cache}^{(i)}$ denote the per-token output cost, standard input cost, and cache-hit input cost of $M_i$, respectively, with $c_\text{cache}^{(i)} < c_\text{in}^{(i)}$.

Suppose at dialogue turn $t$, the previous dialogue history $\tau_{t-1}$, the current user query $x_t$, and the response $y_t$ generated by model $M_{a_t}$ have token counts $L(\tau_{t-1})$, $L(x_t)$, and $L(y_t)$, respectively. The output cost reward is defined as:
\begin{equation}
    R_t^\text{out} = -c_\text{out}^{(a_t)} \cdot L(y_t).
\end{equation}
The input cost accounts for cache hits: if the same model is used in consecutive turns, cached history tokens is charged at the lower cache cost, while the remaining tokens are charged at the standard input cost; if the model switches, all tokens are charged at the standard input cost:
\begin{equation}
\begin{split}
R_t^\text{in} =
& - \Bigl[ 
    c_\text{in}^{(a_t)} \bigl( (L(\tau_{t-1}) \bmod B) + L(x_t) \bigr) \\
& \quad + c_\text{cache}^{(a_t)} \Bigl\lfloor \frac{L(\tau_{t-1})}{B} \Bigr\rfloor \cdot B 
  \Bigr], \quad \text{if } a_t = a_{t-1} \\[0.8ex]
& - c_\text{in}^{(a_t)} \bigl( L(\tau_{t-1}) + L(x_t) \bigr), \quad \text{if } a_t \ne a_{t-1}
\end{split}
\end{equation}

where $B$ is the token size of a KV cache block. The total cost reward is then defined as $R_t^\text{cost} = R_t^\text{in} + R_t^\text{out}$. Accordingly, the original reward function (Eq.~\ref{rewardf}) can be extended as:
\begin{equation}
    \mathcal{R}(s_t, a_t) = R_t - R_{t-1} + \lambda \cdot R_t^\text{cost},
\end{equation}
where $\lambda$ is a hyperparameter controlling the trade-off between dialogue quality and cost. This extension allows MCTS search and routing policy training to jointly consider performance and cost, achieving a balanced trade-off.

\begin{table*}[ht]
\caption{Performance comparison of DialRouter and various baselines across different dialogue tasks using Qwen and Llama candidate sets. Results of stochastic methods (Random, Greedy Router, and MCTS Router) are averaged over three random seeds. The best results among practical methods are shown in \textbf{bold}, and the second-best are \underline{underlined}.}
\centering
\makebox[1\textwidth][c]{
\resizebox{1\textwidth}{!}{
\begin{tabular}{l|ccccccc|ccccccc}
\toprule
& \multicolumn{7}{c}{\textbf{Qwen Series Set}} & \multicolumn{7}{c}{\textbf{Llama Series Set}} \\
\cmidrule(lr){2-8} \cmidrule(lr){9-15}
& \multicolumn{2}{c}{ShareGPT} &\multicolumn{2}{c}{JDDC} & \multicolumn{2}{c}{MedDG} & & \multicolumn{2}{c}{ShareGPT} &\multicolumn{2}{c}{JDDC} & \multicolumn{2}{c}{MedDG} & \\
\cmidrule(lr){2-3} \cmidrule(lr){4-5} \cmidrule(lr){6-7} \cmidrule(lr){9-10} \cmidrule(lr){11-12} \cmidrule(lr){13-14}
\multirow{-4}{*}{Method} & SR$\uparrow$ & AT$\downarrow$ & SR$\uparrow$ & AT$\downarrow$ & SR$\uparrow$ & AT$\downarrow$ & \multirow{-2}{*}{SR$_\text{avg}\uparrow$} & SR$\uparrow$ & AT$\downarrow$ & SR$\uparrow$ & AT$\downarrow$ & SR$\uparrow$ & AT$\downarrow$ & \multirow{-2}{*}{SR$_\text{avg}\uparrow$}\\ \midrule
Qwen3-4B / Llama3.2-1B-Instruct& 
81.75 & 5.09 & 73.45 & 3.82 & 71.04 & 5.03 & 75.41 &
63.03 & 6.62 & 48.07 & 4.43 & 43.61 & 6.97 & 51.57 \\
Qwen3-8B / Llama3.2-3B-Instruct & 
82.80 & 5.03 & 76.14 & 3.50 & 71.82 & 5.02 & 76.92 &
73.77 & 5.92 & 70.65 & 3.94 & 60.78 & 6.44 & 68.40 \\
Qwen3-14B / Llama3.1-8B-Instruct &
83.24 & 5.11 & 77.19 & \textbf{3.36} & 72.50 & 5.27 & 77.64 &
79.24 & \textbf{5.60} & 73.21 & \textbf{3.59} & 65.35 & 6.19 & 72.60 \\ \midrule
Random \cite{lu2024blending} &
81.50 & 4.95 & 76.60 & 3.52 & 71.46 & \underline{4.86} & 76.52 &
76.16 & 6.12 & 68.97 & 4.00 & 61.11 & 6.34 & 68.75 \\
Prompt LLM \cite{feng2025graphrouter} &
82.85 & \textbf{4.92} & 76.49 & 3.50 & \underline{72.74} & 5.29 & 77.36 &
79.19 & 5.79 & 72.49 & 3.91 & 65.35 & 6.19 & 72.34 \\
KNN Router \cite{hu2024routerbench} &
83.37 & 5.01 & \underline{78.44} & \textbf{3.32} & 71.99 & \textbf{4.75} & 77.93 &
\underline{81.71} & 5.77 & \underline{74.82} & 3.72 & 64.58 & \textbf{6.05} & \underline{73.70} \\
MF Router \cite{ong2025routellm} &
82.41 & 5.13 & 76.59 & 3.73 & 72.41 & 4.97 & 77.14 &
80.01 & 5.67 & 72.06 & 3.66 & \underline{66.19} & 6.30 & 72.75 \\
RouterDC \cite{chen2024routerdc} &
82.72 & \underline{4.94} & 77.19 & \underline{3.36} & 72.50 & 5.27 & 77.47 &
80.94 & 5.68 & 73.78 & \underline{3.60} & 66.09 & 6.36 & 73.60 \\
Avengers \cite{zhang2025avengers} &
\underline{83.93} & 5.00 & 78.39 & 3.57 & 71.69 & 5.00 & \underline{78.00} &
80.38 & \underline{5.61} & 73.99 & 3.70 & 66.14 & 6.65 & 73.50 \\ \midrule
\rowcolor{lightpurple} \textbf{DialRouter (Ours)} &
\textbf{87.46} & 4.96 & \textbf{83.45} & 3.47 & \textbf{79.03} & 4.98 & \textbf{83.31} &
\textbf{85.82} & 5.72 & \textbf{80.16} & 3.90 & \textbf{70.65} & \underline{6.16} & \textbf{78.88} \\ \midrule
Greedy Router &
88.50 & 5.12 & 83.60 & 3.54 & 79.71 & 4.95 & 83.94 &
85.75 & 5.73 & 80.83 & 3.73 & 72.51 & 6.30 & 79.70 \\
MCTS Router &
94.11 & 4.97 & 88.42 & 3.55 & 86.77 & 4.77 & 89.77 &
91.94 & 5.76 & 85.28 & 3.86 & 82.38 & 6.41 & 86.53 \\ \bottomrule
\end{tabular}
}}
\label{tab:main}
\vskip-0.1in
\end{table*}
\section{Experiments}
\subsection{Experimental Setup}
\textbf{Datasets.} We conduct experiments on multi-turn dialogue tasks using three real-world datasets, including one open-domain dataset, \textbf{ShareGPT}\footnote{\url{https://sharegpt.com}}, collected from real user–LLM conversations, and two domain-specific datasets, \textbf{JDDC} (Jing Dong Dialogue Corpus) \cite{chen2020jddc}, collected from real-world e-commerce customer-service interactions, and \textbf{MedDG} \cite{liu2022meddg}, collected from real online medical consultation scenarios. Instead of directly using the original dialogue utterances, we extract user profiles and checklists from each dialogue instance to construct user simulators and reward model, thereby enabling interactive environments for corresponding multi-turn dialogue tasks.

To evaluate domain generalization, we hold out a subset of ShareGPT centered on legal and financial topics, forming a domain-generalization test set, denoted as \textbf{ShareGPT-LF}, which is completely non-overlapping with the training data. Further details of the datasets are provided in Appendix~\ref{appena}.

\textbf{Candidate LLM Sets.} We selected mainstream large language models of different series and scales, forming four candidate sets:  (i) \textbf{Qwen series} \cite{yang2025qwen3}: Qwen3-4B, Qwen3-8B, Qwen3-14B;  (ii) \textbf{Llama series} \cite{grattafiori2024llama}: Llama3.2-1B-Instruct, Llama3.2-3B-Instruct, Llama3.1-8B-Instruct;  (iii) \textbf{Mixed series}: a combination of six models from the Qwen and Llama series;  (iv) \textbf{Closed-source LLMs}: Qwen3-Max, DeepSeek-Chat-V3.2, GPT-5.1. Implementation details and pricing information are provided in Appendix~\ref{appenc}; among the closed-source LLMs, DeepSeek is priced significantly lower than the others.

\textbf{Evaluation Metrics.} We use \textbf{Success Rate (SR)} to measure the completion of dialogue tasks. For a single task, SR is defined as the average completion of checklist items at the final dialogue turn $T$:
\begin{equation}
    SR = \frac{1}{n} \sum_{i=1}^{n} r_i^{(T)} = \frac{1}{n} R_T = \frac{1}{n} \sum_{t=1}^{T} \mathcal{R}(s_t, a_t),
\end{equation}
reflecting the cumulative dialogue reward without considering costs, where a higher SR indicates better task performance. \textbf{Average Turns (AT)} measures the mean number of dialogue turns required to complete a task, with lower values being better. For the closed-source LLM, we further introduce \textbf{Cost}, defined as the total monetary expense of model invocations over a dataset based on API pricing rules.

\textbf{Baselines.} We compare DialRouter with a range of baseline methods, including: \textbf{Single Fixed LLM}, which uses a single model for all tasks; \textbf{Random LLM} \cite{lu2024blending}, which selects models randomly; \textbf{Prompt LLM} \cite{feng2025graphrouter}, which leverages meta-prompts for model selection; \textbf{RouterDC} \cite{chen2024routerdc}, which employs contrastive learning to align query and LLM embeddings; \textbf{KNN Router} \cite{hu2024routerbench}, which relies on the historical performance of similar queries; \textbf{Matrix Factorization (MF)} \cite{ong2025routellm, zhuang2025embedllm}, which reconstructs LLM correctness patterns via low-rank latent spaces; and \textbf{Avenger} \cite{zhang2025avengers}, which selects models based on clustering. Additionally, we include two simulation-based routing methods that utilize the user simulator and reward model: \textbf{Greedy Router}, which performs one-step greedy search, and \textbf{MCTS Router}, which conducts MCTS.

\textbf{Implementation Details.} For MCTS, we use $K=10$ simulations, an exploration coefficient $c=2$, and a discount factor $\gamma=0.999$. The maximum dialogue turn is set to $T_{\max}=8$. For retrieval-based future state approximation, we employ BGE-M3 \cite{chen2024m3} as the fixed retriever. The DialRouter is trained with a Qwen2.5-0.5B-based state encoder, where all but the last three transformer layers are frozen to mitigate overfitting. The user simulator and the reward model are implemented using Qwen3-Max. The routing model is optimized using AdamW \cite{loshchilov2017decoupled} with a learning rate of $1\mathrm{e}{-5}$, a batch size of 32, and trained for 15 epochs. All experiments are conducted on a single NVIDIA RTX 4090D GPU.

\subsection{Main Results}
\textbf{Oracle Analysis.} The results in Table~\ref{tab:main} reveal a clear tiered performance pattern, characterized by a large performance gap between the MCTS and the Greedy Router, as well as a consistent gap between the Greedy Router and the strongest single LLMs. Although simulation-based routing is impractical at inference time, the MCTS Router provides a search-based reference that serves as an approximate, oracle-like upper bound on achievable performance under the current candidate set. Notably, the consistent gap between MCTS and the Greedy Router further highlights the critical role of long-term awareness in multi-turn dialogue routing.

\textbf{Comparison with Single LLMs.} Table~\ref{tab:main} reports the performance of DialRouter on multi-turn dialogue tasks over the Qwen and Llama candidate sets. DialRouter achieves an average success rate of 83.31\% on the Qwen series, surpassing Qwen3-14B by 5.67\%, and 78.88\% on the Llama series, exceeding Llama3.1-8B-Instruct by 6.28\%, consistently outperforming the strongest single LLM in each set. These results indicate that DialRouter can effectively exploit the complementary strengths of heterogeneous models, thereby breaking the performance ceiling of individual LLMs.

\textbf{Comparison with Baseline Routers.} Compared with representative routing baselines, DialRouter improves the average success rate by more than 5\% over KNN Router, Avengers, and RouterDC across both candidate sets, and achieves overall performance comparable to the simulation-based Greedy Router. This aligns with our motivation that long-term awareness is crucial for effective multi-turn dialogue routing beyond myopic decisions, showing that learning a router grounded in long-horizon planning enables high-quality routing without requiring online search. Although the average turns slightly increases in some scenarios, DialRouter maintains consistently higher task success rates.

\begin{table}[h]
\caption{Performance comparison of DialRouter and baselines on ShareGPT using mixed candidate set of Qwen and Llama models.}
\centering
\small
\begin{tabular}{l|cc}
\toprule
Method & SR$\uparrow$ & AT$\downarrow$ \\ \midrule
Random \cite{lu2024blending} & 79.50 & 5.74 \\
KNN Router \cite{hu2024routerbench} & 84.23 & 5.22 \\
Avengers \cite{zhang2025avengers} & 84.72 & \textbf{5.12} \\ \midrule
\textbf{DialRouter (Ours)} & \textbf{88.32} & 5.21\\ \midrule
Greedy Router & 90.06 & 5.17 \\
MCTS Router & 94.99 & 5.13 \\ \bottomrule
\end{tabular}
\label{tab:mix}
\end{table}
\textbf{Adaptability on Mixed Series Candidate Sets.} To evaluate the generalization and decision robustness of DialRouter when the candidate set includes multiple LLM families, we conducted experiments on ShareGPT with the Mixed series, which combines the Qwen and Llama series. As shown in Table~\ref{tab:mix}, DialRouter achieves an SR of 88.32\%, surpassing the baseline routers by over 3.60\%, demonstrating its strong adaptability in selecting among heterogeneous LLM architectures. Notably, DialRouter performs even better on the mixed candidate set than on single-series candidate sets, indicating its ability to cross LLM families and effectively identify and integrate the complementary strengths of different LLMs within a larger candidate pool.

\textbf{Cross-Domain Generalization.} To evaluate the cross-domain generalization of DialRouter, we conduct experiments on ShareGPT-LF. As shown in Table~\ref{tab:lf}, all methods are trained on ShareGPT, which excludes legal and financial topics. DialRouter achieves success rate improvements of 1.95\%–3.45\% on the Qwen series and 1.56\%–2.85\% on the Llama series over the best single LLM and other baselines. These results demonstrate that DialRouter maintains a performance advantage in unseen domains, indicating that the learned routing policy can effectively leverage LLM complementarity to achieve robust cross-domain generalization.
\begin{table}[t]
\caption{Cross-domain generalization results on ShareGPT-LF.}
\centering
\makebox[1\columnwidth][c]{
\resizebox{1\columnwidth}{!}{
\begin{tabular}{l|cccc}
\toprule
& \multicolumn{2}{c}{Qwen Series Set} & \multicolumn{2}{c}{Llama Series Set} \\
\cmidrule(lr){2-3} \cmidrule(lr){4-5}
\multirow{-2}{*}{Method} & SR$\uparrow$ & AT$\downarrow$ & SR$\uparrow$ & AT$\downarrow$\\ \midrule
Qwen3-4B / Llama3.2-1B-Instruct &
80.24 & 4.80 & 62.25 & 6.44\\
Qwen3-8B / Llama3.2-3B-Instruct &
80.66 & 4.68 & 75.17 & 5.36\\
Qwen3-14B / Llama3.1-8B-Instruct &
81.74 & 4.65 & 80.18 & \textbf{5.09}\\ \midrule
Random \cite{lu2024blending} & 
81.03 & 4.67 & 77.62 & 5.76 \\
KNN Router \cite{hu2024routerbench} &
81.64 & \textbf{4.56} & 81.47 & 5.15 \\
Avengers \cite{zhang2025avengers} &
81.30 & 4.67 & 80.18 & \textbf{5.09} \\ \midrule
\textbf{DialRouter (Ours)} & 
\textbf{83.69} & 4.64 & \textbf{83.03} & 5.48\\ \midrule
Greedy Router & 
87.18 & 4.53 & 86.47 & 5.26\\
MCTS Router & 
91.46 & 4.50 & 90.07 & 5.14\\ \bottomrule
\end{tabular}
}}
\label{tab:lf}
\vskip-0.1in
\end{table}

\subsection{Results of Cost-aware Routing}
\begin{table}[h]
\caption{Performance and cost comparison of DialRouter and baselines on ShareGPT using Closed-source model candidate set. Cost-aware rewards are applied in the Balance setting. Cost (\$) refer to the total monetary cost for the test set.}
\centering
\makebox[1\columnwidth][c]{
\resizebox{1\columnwidth}{!}{
\begin{tabular}{l|ccc|ccc}
\toprule
 & \multicolumn{3}{c}{Performance First} & \multicolumn{3}{c}{Balance} \\
 \cmidrule(lr){2-4} \cmidrule(lr){5-7}
\multirow{-2}{*}{Method} & SR$\uparrow$ & AT$\downarrow$ & Cost$\downarrow$ & SR$\uparrow$ & AT$\downarrow$ & Cost$\downarrow$ \\ \midrule
Qwen3-Max & 
84.09 & \textbf{4.53} & 0.30 & 84.09 & \textbf{4.53} & 0.30 \\
DeepSeek-Chat-V3.2 & 
85.13 & 4.66 & \textbf{0.04} & 85.13 & 4.66 & \textbf{0.04} \\
GPT-5.1 & 
84.57 & 4.80 & 1.05 & 84.57 & 4.80 & 1.05\\ \midrule
Random \cite{lu2024blending} & 
84.41 & 4.57 & 0.52 & 84.41 & 4.57 & 0.52\\
KNN Router \cite{hu2024routerbench} &  
85.91 & 4.56 & 0.50 & 85.13 & 4.66 & \textbf{0.04}\\
Avengers \cite{zhang2025avengers} &  
84.54 & 4.56 & 0.32 & 85.13 & 4.66 & \textbf{0.04}\\ \midrule
\textbf{DialRouter (Ours)} & 
\textbf{89.92} & 4.59 & 0.44 & \textbf{88.69} & 4.63 & 0.09\\ \midrule
Greedy Router &  
91.80 & 4.66 & 0.53 & 91.28 & 4.43 & 0.20\\
MCTS Router &  
96.11 & 4.52 & 0.46 & 95.93 & 4.44 & 0.19\\ \bottomrule
\end{tabular}
}}
\label{tab:cost}
\end{table}
\textbf{Overall Performance under Cost-aware Routing.} Table~\ref{tab:cost} reports the results of DialRouter on the closed-source LLM candidate set with cost reward on ShareGPT. Under the Performance First setting ($\lambda=0$), DialRouter achieves the highest success rate of 89.92\%, substantially outperforming strong single closed-source LLMs such as DeepSeek-Chat-V3.2 (+4.97\%) and GPT-5.1 (+5.35\%), as well as all routing baselines. Under the Balance setting ($\lambda=50$), incorporating cost reward reduces the average cost from 0.44 to 0.09, corresponding to an approximately 80\% reduction, while the success rate only decreases to 88.69\%, amounting to a relative drop of about 1.4\%. In contrast, several baselines exhibit clear policy degradation, effectively collapsing to a single low-cost DeepSeek selection, thereby abandoning the decision-making role expected of a router. These results demonstrate that, with cost reward, DialRouter achieves a more balanced performance-cost trade-off than baseline methods, substantially reducing inference cost with only a limited degradation in task performance.

\begin{figure}[t]
    \centering
    \includegraphics[width=1.0\linewidth]{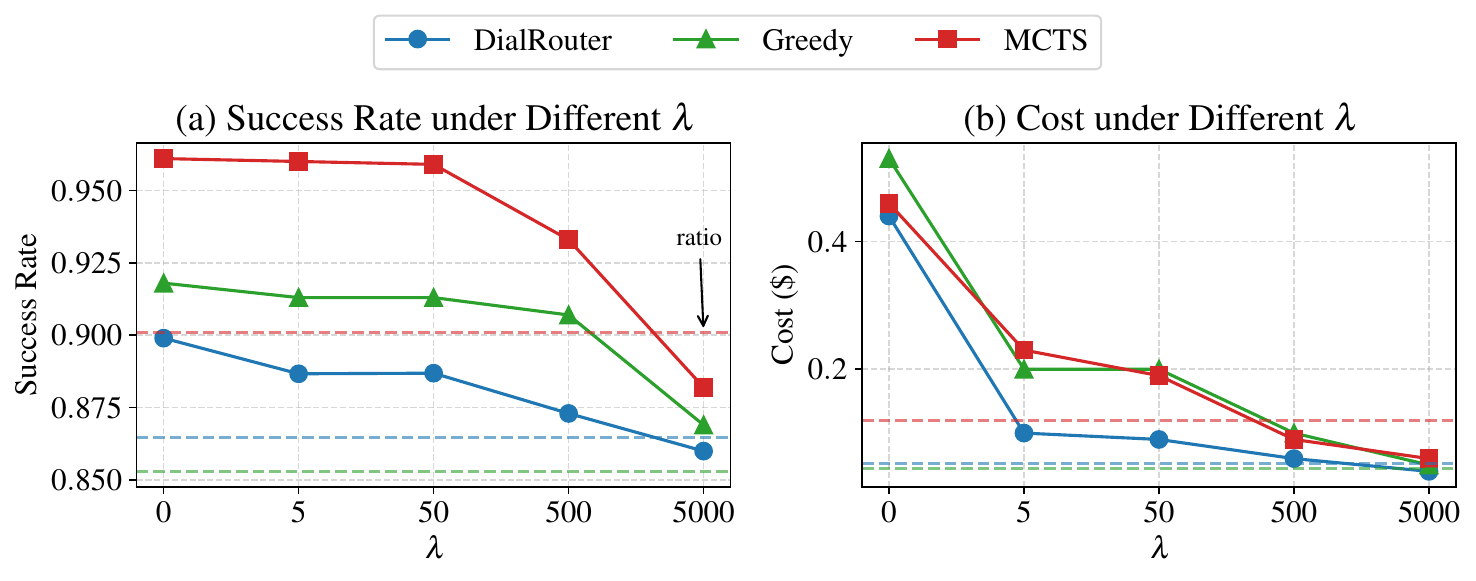}
    \caption{SR and cost of DialRouter, Greedy Router, and MCTS Router under different values of the cost weight $\lambda$. Dashed horizontal lines correspond to the ratio-based reward setting.}
    \label{fig:l}
    \vskip-0.1in
\end{figure}
\textbf{Cost and Performance under Varying Cost Weights $\lambda$.} As shown in Figure~\ref{fig:l}, we analyze the SR and cost of DialRouter, Greedy Router, and MCTS Router under different $\lambda$. Overall, increasing $\lambda$ consistently reduces cost for all methods, accompanied by a moderate decline in SR. When $\lambda$ increases from 0 to 5, cost is substantially reduced with only a slight SR drop, indicating that a small $\lambda$ can effectively lower inference cost with minimal performance impact. As $\lambda$ increases from 5 to 50, both SR and cost become relatively insensitive to $\lambda$, corresponding to a stable performance-cost trade-off regime. Across all $\lambda$, DialRouter consistently incurs lower cost than MCTS, which can be attributed to behavior cloning bias toward high-frequency low-cost actions in MCTS trajectories, underrepresenting infrequent yet performance-critical high-cost decisions; this results in lower overall cost with a SR degradation. We further evaluate a ratio-based reward $\mathcal{R} = \frac{R_t - R_{t-1}}{R_t^\text{cost}}$, shown as dashed lines in Figure~\ref{fig:l}; under this formulation, Greedy Router almost always selects DeepSeek due to its dominant low price, while both MCTS Router and DialRouter also exhibit performance degradation, indicating that ratio-based rewards fail to stably capture the performance--cost trade-off in candidate sets with substantial price disparities.

\subsection{Ablation Study}
\begin{table}[t]
\caption{Ablation study of DialRouter components, showing the effects of removing MCTS-based policy discovery, the retrieval module, and using alternative information fusion strategies across three dialogue datasets using the Qwen candidate set.}
\centering
\makebox[1\columnwidth][c]{
\resizebox{1\columnwidth}{!}{
\begin{tabular}{l|ccccccc}
\toprule
& \multicolumn{2}{c}{ShareGPT} &\multicolumn{2}{c}{JDDC} & \multicolumn{2}{c}{MedDG} & \\
\cmidrule(lr){2-3} \cmidrule(lr){4-5} \cmidrule(lr){6-7} 
\multirow{-2}{*}{Method} & SR$\uparrow$ & AT$\downarrow$ & SR$\uparrow$ & AT$\downarrow$ & SR$\uparrow$ & AT$\downarrow$ & \multirow{-2}{*}{SR$_\text{avg}\uparrow$} \\ \midrule
\textbf{DialRouter (Ours)} &
87.46 & 4.96 & 83.45 & 3.47 & 79.03 & 4.98 & 83.31\\
\midrule
w/o MCTS & 
83.64 & 4.99 & 80.74 & 3.63 & 74.93 & 4.94 & 79.77\\ \midrule
w/o Ret&
84.78 & 5.09 & 80.37 & 3.64 & 74.92 & 4.94 & 80.02\\
Random Ret&
83.99 & 5.10 & 79.87 & 3.49 & 73.06 & 5.04 & 78.97\\ \midrule
DialRouter (Add Fus.)&
85.27 & 5.21 & 81.35 & 3.42 & 75.92 & 4.80 & 80.85\\ 
DialRouter (Concat Fus.)&
84.85 & 5.02 & 81.40 & 3.55 & 76.05 & 4.93 & 80.77\\ \bottomrule
\end{tabular}
}}
\label{tab:ab}
\vskip-0.1in
\end{table}
\textbf{Effect of MCTS-based Data.} As shown in Table~\ref{tab:ab}, removing MCTS-based data (w/o MCTS) and training the router solely with one-step greedy decisions results in an average SR drop of 3.54\%. This shows that relying only on myopic routing decisions is insufficient to capture long-term effects in multi-turn dialogues, whereas MCTS-generated data provide global, long-term-aware supervision critical to DialRouter’s performance gains.

\textbf{Effect of Retrieval-based Future State Approximation.} Table~\ref{tab:ab} also shows that removing the retrieval module (w/o Ret) or using random retrieval (Random Ret) reduces the average SR by 3.29\% and 4.34\%, respectively, both significantly lower than the complete model. The larger performance drop with random retrieval is primarily due to the introduction of irrelevant information, which amplifies noise and further impairs routing decisions. In contrast, semantically relevant retrieved future states provide more targeted cues, helping the router capture potential dialogue evolution trends and mitigating the difficulty of making long-term decisions based solely on the current state, while incurring only a negligible retrieval overhead of 0.01s per decision.

\textbf{Effect of Fusion Strategies for Future States.} As shown in Table~\ref{tab:ab}, simply incorporating future states without differentiation (Concat Fusion or Add Fusion) yields some improvement but still underperforms the full model with gating. This suggests that state information from different sources varies in reliability and relevance, and a learnable gating mechanism to adaptively fuse them helps suppress retrieval noise and balance current versus future approximated information, resulting in more robust routing decisions.

\section{Conclusion}
In this paper, we present the first systematic study of LLM routing in multi-turn dialogue settings and propose DialRouter, a long-term-aware routing approach for multi-turn dialogue. DialRouter learns routing policies from MCTS-based data, enabling long-horizon routing decisions without online search, and further incorporates retrieval-based approximate future states to assist LLM selection. Experiments across diverse multi-turn dialogue tasks demonstrate that DialRouter effectively integrates the complementary strengths of multiple LLMs, significantly outperforming the strongest single LLMs and existing routing baselines based on local rewards in terms of task success rate; moreover, when augmented with cost-aware rewards, DialRouter achieves a more favorable performance-cost trade-off.
\section*{Impact Statement}
This paper proposes a long-horizon LLM routing method for multi-turn dialogue, which dynamically selects appropriate models based on the dialogue state to optimize long-term conversational objectives without relying on online search, while effectively reducing inference cost without compromising response quality. The broader impact of this work lies in improving user experience and service quality in multi-turn LLM applications, such as customer service, medical consultation. For service providers deploying multiple models, the proposed approach enables more efficient model coordination, thereby enhancing the usability and sustainability of large language model systems.

\bibliography{example_paper}
\bibliographystyle{icml2026}

\newpage
\appendix
\onecolumn

\section{Details of MCTS}\label{append}
In this section, we describe the details of Monte Carlo Tree Search (MCTS) \cite{coulom2006efficient, silver2017mastering}. Starting from the current dialogue state $s_{\text{root}}$ as the root node, MCTS repeatedly performs four phases—Selection, Expansion, Simulation, and Backpropagation—for a total of $K$ iterations, progressively exploring dialogue trajectories induced by different model selections and estimating their long-term returns.

\textbf{Selection.} Starting from the root node $s_{\text{root}}$, MCTS recursively selects child nodes until reaching a leaf node that is not fully expanded. During this process, we adopt the Upper Confidence Tree (UCT) \cite{silver2017mastering} criterion to balance exploration and exploitation:
\begin{equation}
    UCT(s,a) = Q(s,a) + c \sqrt{\frac{\ln N(s)}{N(s,a) + 1}},
\end{equation}
where $Q(s,a)$ denotes the estimated cumulative return of taking action $a$ at state $s$, $N(s)$ is the number of visits to state $s$, $N(s,a)$ is the number of times action $a$ has been selected at state $s$, and $c$ is a constant controlling the degree of exploration. At each step, the action maximizing $UCT(s,a)$ is selected.

\textbf{Expansion.} Upon reaching a leaf node, MCTS selects an unexplored action, corresponding to a model routing decision for the next dialogue turn, and generates a new dialogue state, which is then added to the search tree.

\textbf{Simulation.} From the newly expanded node, MCTS performs an approximate rollout to estimate its potential return. Specifically, we adopt a one-step greedy routing strategy, selecting at each turn the currently estimated optimal model, and advance the dialogue using the user simulator until termination (e.g., when the user simulator outputs a termination signal or the maximum dialogue length $T_{\max}$ is reached).

\textbf{Backpropagation.} After a simulation terminates, the accumulated return is propagated backward along the simulated trajectory from the leaf node to the root. For each state $s_t$ on the trajectory, the return is defined as the discounted cumulative reward from that state until the end of the dialogue:
\begin{equation}
    V(s_t) = \sum_{k=t}^{T} \gamma^{k-t} \mathcal{R}(s_k, a_k),
\end{equation}
where $\gamma \in (0,1]$ is the discount factor. Subsequently, for each encountered state--action pair $(s,a)$, the visit count and value estimate are updated as:
\begin{equation}
\begin{aligned}
N(s,a) &\leftarrow N(s,a) + 1, \\
Q(s,a) &\leftarrow Q(s,a) + \frac{V(s) - Q(s,a)}{N(s,a)}.
\end{aligned}
\end{equation}
After completing all $K$ search iterations, we select, at the root node $s_{\text{root}}$, the action with the highest estimated value as the optimal model choice for the current state:
\begin{equation}
    a^* = \arg\max_a Q(s_{\text{root}}, a).
\end{equation}
\section{Details of Datasets}\label{appena}
\subsection{Dataset Sources and Characteristics}
\textbf{ShareGPT}\footnote{\url{https://sharegpt.com}} is an English open-domain multi-turn dialogue dataset collected from real user–LLM interactions, where users voluntarily share conversation links generated on the ChatGPT web interface. The dataset covers a wide range of general-purpose tasks, including information seeking, writing assistance, code generation, and daily conversations, reflecting diverse real-world usage scenarios of conversational AI systems.

\textbf{JDDC (Jing Dong Dialogue Corpus)} \cite{chen2020jddc} is a Chinese e-commerce customer-service multi-turn dialogue dataset released by JD.com, derived from real interaction logs between users and customer-service systems. The dataset focuses on task-oriented dialogue scenarios in the e-commerce domain, covering typical business processes such as product inquiry, order management, logistics, and after-sales support.

\textbf{MedDG} \cite{liu2022meddg} is a Chinese medical multi-turn dialogue dataset collected from real online health consultation communities, recording multi-turn interactions between users and medical professionals or consultation systems. The dataset centers on medical consultation scenarios and includes tasks such as symptom description, disease diagnosis, and health advice, exhibiting strong domain specificity and long-term information dependencies.

\subsection{Data Splits}
\begin{table}[h]
\caption{Statistics of dialogue-level data splits. All splits are performed at the dialogue level.}
\centering
\small
\begin{tabular}{lccc}
\toprule
Dataset & \# Training Dialogues & \# Test Dialogues & Avg. \# Checklist Items \\ \midrule
ShareGPT & 750 & 90 & 6.70\\
JDDC & 750 & 100 & 4.99\\
MedDG & 750 & 100 & 4.94\\
ShareGPT-LF & - & 66 & 6.58\\\bottomrule
\end{tabular}
\label{tab:data}
\end{table}
As shown in Table~\ref{tab:data}, for JDDC and MedDG, we randomly sample dialogue instances and split them into training and test sets. For ShareGPT, to ensure consistent intent decomposition and checklist-based evaluation, we apply a data filtering step that retains multi-turn dialogues with moderate lengths and coherent intent flows, while removing instances involving sensitive or unsafe content that may trigger API-level safety refusals. Based on the filtered ShareGPT data, we further construct ShareGPT-LF by isolating dialogues from the legal and financial domains to evaluate cross-domain generalization, and randomly split the remaining data into training and test sets.

\subsection{Topic Distribution of ShareGPT}
\begin{table}[h]
\caption{Coarse-grained topic distribution of the ShareGPT dataset.}
\centering
\small
\begin{tabular}{lcccccccc}
\toprule
Topic & Programming & Science & Health & Technology & Culture & Law & Finance & Others\\
\midrule
Train & 419 & 187 & 23 & 23 & 20 & 0 & 0 & 78\\
Test & 48 & 21 & 6 & 4 & 2 & 0 & 0 & 9\\
SharGPT-LF & 0 & 0 & 0 & 0 & 0 & 22 & 44 & 0 \\
\bottomrule
\end{tabular}
\label{tab:sharegpt_topics}
\end{table}
We analyze the topic distribution of ShareGPT using coarse-grained categories. As shown by Tabel~\ref{tab:sharegpt_topics}, the dataset is dominated by programming and science related dialogues, while also covering a diverse range of topics such as health, technology, and culture. The topic distributions of the training and test splits are broadly consistent.
\subsection{User Profiling and Checklist Construction}\label{appena3}
For each dialogue instance in the dataset, we prompt Qwen3-Max to extract a structured user profile and a corresponding checklist. Representative examples are shown in Tables~\ref{tab:sg},~\ref{tab:jddc}, and~\ref{tab:meddg}. The extracted user profiles and checklists are designed to capture dialogue-level user intent and evaluation criteria, while allowing for domain-specific variations across datasets.

For \textbf{ShareGPT}, the user profile represents an intent chain with explicit triggering logic, capturing the deterministic progression of user interests, while the checklist focuses on content correctness and coverage, adherence to formatting and instruction constraints, and overall dialogue coherence. For \textbf{JDDC}, the user profile includes user behavior patterns, contextual background, and explicit goals in the customer-service setting, and the checklist emphasizes effective user guidance, correctness of information delivery, and high-quality emotional support and need fulfillment. For \textbf{MedDG}, the user profile captures the chief complaint, detailed symptom descriptions, treatment history, and basic consultation goals, whereas the checklist evaluates proper information elicitation and clarification, correctness of medication and examination recommendations, accuracy of disease assessment, and the provision of appropriate health advice and patient reassurance.

We formalize each dialogue task as a tuple $(p_{\mathrm{usr}}, C, x_1)$, where $p_{\mathrm{usr}}$ denotes the extracted user profile, $C$ the checklist, and $x_1$ the initial user query of the dialogue instance. For JDDC, we additionally extract domain-specific customer-service prior knowledge and provide it to all candidate models to ensure alignment on required business rules.
\begin{table}[H]
\caption{An example dialogue from ShareGPT used for user profiling and checklist construction. Some content is omitted for brevity.}
\centering
\begin{tabular}{ll}
\toprule
\textbf{Category} & \textbf{Content Details}\\\midrule
\textbf{Dialogue Instance} &
\parbox[t]{13.5cm}{
\textbf{User}: What are the classic cocktails that every bartender should know how to make and what are their ingredients?\\
\textbf{Assistant}: 1. Old Fashioned: Ingredients - bourbon or rye whiskey, sugar, bitters, ice, orange twist or cherry for garnish. 2. Martini: Ingredients - gin or vodka, dry vermouth, ice, olives or lemon twist ...\\
\textbf{User}: I've had all of these classic cocktails, but I'm looking for something new to try. Do you have any recommendations?\\
\textbf{Assistant}: Sure, here are a few modern cocktail recommendations to try...\\
\textbf{User}: The Elderflower Sour sounds interesting. I love trying new cocktails with unique ingredients.\\
\textbf{Assistant}: The Elderflower Sour is a great cocktail...\\
\textbf{User}: Definitely going to try the Elderflower Sour tonight! Do you have any other unique cocktail recommendations with interesting flavors? I'm in the mood to experiment.\\
\textbf{Assistant}: Sure, here are a few more unique cocktail recommendations...\\
\textbf{User}: The Smoked Maple Old Fashioned sounds intriguing. How exactly do you use a smoking gun in a cocktail?\\
\textbf{Assistant}: To use a smoking gun in a cocktail like the Smoked Maple Old Fashioned...\\
\textbf{User}: Wow, I never knew you could use a smoking gun in cocktails. That's a really cool technique! Do you have any other creative cocktail techniques that I should know about?\\
\textbf{Assistant}: Yes, there are several creative cocktail techniques...\\
\textbf{User}: Wow, these techniques are really cool! I've never heard of fat-washing before. Do you have any tips for getting the proportions right?\\
\textbf{Assistant}: Fat-washing is a fun technique...
}
\\ \midrule
\textbf{User Profile} & 
\parbox[t]{13.5cm}{
\textbf{User Intent Chain with Trigger Logic}:\\
(I1) Request classic cocktails with ingredients and preparation methods; triggered initially when the user seeks foundational cocktail knowledge and re-triggered if recipes are insufficient.\\
(I2) Request modern cocktail recommendations with recipes; triggered after (I1) is fulfilled and re-triggered if preparation details are incomplete.\\
(I3) Request unique or creative cocktail recipes with distinctive flavors; triggered when the user expresses interest in experimentation and re-triggered if diversity is lacking.\\
(I4) Request creative cocktail-making techniques; triggered when the user shows interest in how creative cocktails are prepared and re-triggered if key techniques are omitted.
}
\\ \midrule
\textbf{Checklist} & 
\parbox[t]{13.5cm}{
(1) Whether the model’s recommended classic cocktails include Old Fashioned and Bloody Mary.\\
(2) Whether the model’s recommended modern cocktails include Elderflower Sour.\\
(3) Whether the model’s output includes at least 5 classic cocktails.\\
(4) Whether the model’s output includes at least 5 modern cocktails.\\
(5) Whether the model provides the recipe and basic preparation method for each cocktail.\\
(6) Whether the model avoids repeating cocktails or other information already provided.\\
(7) Whether the model avoids self-contradictions across multiple turns.
} 
\\
\bottomrule
\end{tabular}
\label{tab:sg}
\end{table}

\begin{table}[H]
\caption{An example dialogue from JDDC used for user profiling and checklist construction. The original dialogue, user profile, and checklist are in Chinese and are translated into English for presentation, with some content omitted for brevity.}
\centering
\begin{tabular}{ll}
\toprule
\textbf{Category} & \textbf{Content Details}\\\midrule
\textbf{Dialogue Instance} &
\parbox[t]{13.5cm}{
\textbf{User}: Can you help me change the delivery address for my order?\\
\begin{CJK}{UTF8}{gbsn}
(可以帮我改下订单的地址吗？)
\end{CJK}\\
\textbf{Customer Service}: If the address is within the same city, you can contact the courier to update it.\\
\begin{CJK}{UTF8}{gbsn}
(同一市内可以联系配送员直接修改的哦。)
\end{CJK}\\
\textbf{User}: It’s not the same city. The current address is in Shanghai, but I’ll be returning to Anhui tomorrow.\\
\begin{CJK}{UTF8}{gbsn}
(不在同一个城市，现在地址是上海，但是我明天要回安徽。)
\end{CJK}\\
\textbf{Customer Service}: Sorry, cross-city address changes are not supported. We recommend placing a new order instead.\\
\begin{CJK}{UTF8}{gbsn}
(抱歉，地址在不同城市不能操作的，只能建议您重新下单哦。)
\end{CJK}\\
\textbf{User}: If I cancel the order, how long will it take to receive the refund?\\
\begin{CJK}{UTF8}{gbsn}
(那我取消订单的话退款多久到账呢？)
\end{CJK}\\
\textbf{Customer Service}: Refund times depend on the payment method: one business day for digital wallets, 1--7 business days for debit cards, and 1--15 business days for credit cards.\\
\begin{CJK}{UTF8}{gbsn}
(微信零钱1个工作日内到账，储蓄卡1-7个工作日内到账，信用卡1-15个工作日内到账的哦！)
\end{CJK}\\
\textbf{User}: Why can’t the address be changed? This is really inconvenient.\\
\begin{CJK}{UTF8}{gbsn}
(为什么不能改地址，你们这也太不方便了。)
\end{CJK}\\
\textbf{Customer Service}: We apologize for the inconvenience. Our logistics system still has limitations.\\
\begin{CJK}{UTF8}{gbsn}
(非常抱歉，我们物流还有待完善呢。)
\end{CJK}\\
\textbf{User}: This is quite troublesome---I need the item urgently.\\
\begin{CJK}{UTF8}{gbsn}
(这也太麻烦了，我还急着用呢。)
\end{CJK}\\
\textbf{Customer Service}: We understand your urgency and will continue to improve our service.\\
\begin{CJK}{UTF8}{gbsn}
(非常抱歉！如果是我的话我也会很着急的，我们会改进的！)
\end{CJK}\\
\textbf{User}: Alright.\\
\begin{CJK}{UTF8}{gbsn}
(行吧。)
\end{CJK}\\
\textbf{Customer Service}: Thank you for your understanding. Is there anything else we can help with?\\
\begin{CJK}{UTF8}{gbsn}
(谢谢您的理解！还有什么能帮到您的吗？)
\end{CJK}
}
\\ \midrule
\textbf{User Profile} & 
\parbox[t]{13.5cm}{
\textbf{Behavior Patterns}: Attempts to change the delivery address across cities; expresses urgency and dissatisfaction; proactively inquires about refund timing.\\
\textbf{General Context}: Currently in Shanghai and returning to Anhui the next day; has an urgent need for the item.\\
\textbf{User Goals}:\\
(1) Change the delivery address from Shanghai to Anhui.\\
(2) Inquire about the refund processing time if the order is canceled.
}
\\ \midrule
\textbf{Checklist} & 
\parbox[t]{13.5cm}{
(1) Whether the agent identifies or confirms the location involved in the address change request.\\
(2) Whether the agent clearly explains that delivery addresses can only be modified within the same city and explicitly states that cross-city address changes are not supported.\\
(3) Whether the agent acknowledges and soothes the user's urgency and dissatisfaction.\\
(4) Whether the agent suggests placing a new order as an alternative solution.\\
(5) Whether the agent provides accurate refund processing times for different payment methods.
} 
\\
\bottomrule
\end{tabular}
\label{tab:jddc}
\end{table}

\begin{table}[H]
\caption{An example dialogue from MedDG used for user profiling and checklist construction. The original dialogue, user profile, and checklist are in Chinese and are translated into English for presentation, with some content omitted for brevity.}
\centering
\begin{tabular}{ll}
\toprule
\textbf{Category} & \textbf{Content Details}\\\midrule
\textbf{Dialogue Instance} &
\parbox[t]{13.5cm}{
\textbf{User}: Since yesterday, I have been experiencing persistent abdominal pain and intermittent diarrhea, about 7--8 times. Last night my temperature was 37.3$^\circ$C. What medication should I take? (Female, 30 years old)\\
\begin{CJK}{UTF8}{gbsn}
(从昨天开始到今天肚子一直痛，断断续续拉了7.8次大便，昨晚测量体温37.3度，请问医生我该吃点什么药?（女，30岁）)
\end{CJK}\\
\textbf{Doctor}: Have you eaten any contaminated food, greasy or spicy food, or consumed alcohol? Do you smoke, feel stressed, or have poor sleep? Have you undergone any medical examinations or taken any medication? Acute gastroenteritis is a possible cause. You may take montmorillonite powder and levofloxacin capsules.\\
\begin{CJK}{UTF8}{gbsn}
(你好，请问你是吃不干净的东西了吗？有没有油腻食物，辛辣刺激性食物，喝酒抽烟，精神压力大，睡眠不好之类有没有做过什么其他检查，有吃过什么药物吗。考虑急性肠胃炎可能性比较大。可以吃蒙脱石散，左氧氟沙星胶囊。)
\end{CJK}\\
\textbf{User}: I ate some greasy food a few days ago. I have not had any examinations recently and have not taken any medication.\\
\begin{CJK}{UTF8}{gbsn}
(前几天吃了点油腻的食物，近期也没有做什么检查，也没有吃什么药。)
\end{CJK}\\
\textbf{Doctor}: That is fine. Both medications are available at pharmacies. I recommend having a stool routine test.\\
\begin{CJK}{UTF8}{gbsn}
(嗯呢。可以的。这两种药药店都有。建议查一个大便常规。)
\end{CJK}\\
\textbf{User}: Okay, thank you.\\
\begin{CJK}{UTF8}{gbsn}
(好的，谢谢。)
\end{CJK}\\
\textbf{Doctor}: Good health relies on self-care rather than only doctors or medications. Sunlight, a positive mindset, and reducing stress are also important.\\
\begin{CJK}{UTF8}{gbsn}
(健康，是靠自己的养护，不是靠医生，和药物，阳光，微笑，就是健康，别给自己压力太大，加油哦。)
\end{CJK}\\
\textbf{User}: Hello doctor, my spouse still has diarrhea after taking the medication, with abdominal pain and diarrhea shortly after meals. Should we continue the medication and observe?\\
\begin{CJK}{UTF8}{gbsn}
("医生你好，我爱人吃了药后，还是拉稀，和刚吃完饭后，肚子痛，又闹肚子，这个还要接着吃药，看看吗。)
\end{CJK}\\
\textbf{Doctor}: Have you had a stool test or blood test? I suggest going to the emergency department for further evaluation. We need to check for possible electrolyte imbalance and abnormal blood counts. Further treatment may be required, and it is difficult to assess without examinations.\\
\begin{CJK}{UTF8}{gbsn}
(有没有做大便检查。抽血了吗。建议你去急诊看看。观察有没有电解质紊乱和血象增高。是需要进一步治疗的。不化验不好说。)
\end{CJK}
}
\\ \midrule
\textbf{User Profile} & 
\parbox[t]{13.5cm}{
\textbf{Demographics}: Female, 30 years old.\\
\textbf{Chief Complaint}: Abdominal pain and diarrhea.\\
\textbf{Symptom Details}:\\
(1) \textit{Abdominal pain}: Located in the lower abdomen; cramping in nature.\\
(2) \textit{Diarrhea}: 7--8 episodes; watery stool; no blood observed.\\
(3) \textit{Fever}: Body temperature of 37.3$^\circ$C; duration within one day.\\
\textbf{Suspected Trigger}: Consumption of greasy food a few days prior.\\
\textbf{Medical History}: None reported.\\
\textbf{Current Medication}: None.\\
\textbf{User Goals}: (1) Symptom assessment; (2) Medication recommendation; (3) Examination recommendation; (4) Diagnosis clarification.
}
\\ \midrule
\textbf{Checklist} & 
\parbox[t]{13.5cm}{
(1) Whether the doctor identifies the patient’s key symptoms, including abdominal pain, diarrhea, and fever.\\
(2) Whether the doctor recommends appropriate medications, such as montmorillonite powder and levofloxacin capsules, for treating diarrhea and abdominal pain.\\
(3) Whether the doctor advises relevant examinations, such as a stool routine test.\\
(4) Whether the doctor determines the underlying condition to be acute gastroenteritis.
} 
\\
\bottomrule
\end{tabular}
\label{tab:meddg}
\end{table}
\section{Details of Candidate LLMs}\label{appenc}
\subsection{Implementation of Candidate LLMs}
For open-source models, we obtain the model weights from the HuggingFace\footnote{\url{https://huggingface.co}} and deploy them locally using the vLLM framework \cite{kwon2023efficient}. For closed-source models, we invoke the APIs of Qwen3-Max\footnote{\url{https://bailian.console.aliyun.com}}, DeepSeek-Chat-V3.2\footnote{\url{https://platform.deepseek.com}}, and GPT-5.1\footnote{\url{https://platform.openai.com/docs/models/gpt-5.1}} through their respective official platforms. Since our study focuses on dialogue flow control and task fulfillment rather than complex long-chain reasoning, all candidate models are configured in non-reasoning mode, with the generation temperature set to 0 to reduce randomness and ensure stable and comparable responses across models.
\subsection{Pricing of Closed-Source LLMs}
\begin{table}[h]
\caption{Pricing of closed-source language models.}
\centering
\small
\begin{tabular}{lccc}
\toprule
Model & Input (\$/1M tokens) & Cached Input (\$/1M tokens)  & Output (\$/1M tokens) \\\midrule
Qwen3-Max & 0.459 & 0.0918 & 1.836 \\
DeepSeek-Chat-V3.2 & 0.28 & 0.028 & 0.42 \\
GPT-5.1 & 1.25 & 0.125 & 10  \\\bottomrule
\end{tabular}
\label{tab:cm}
\end{table}
Pricing information for closed-source models is obtained from the official billing specifications\footnote{\url{https://www.alibabacloud.com/help/en/model-studio/model-pricing}}~\footnote{\url{https://api-docs.deepseek.com/quick_start/pricing}}~\footnote{\url{https://platform.openai.com/docs/pricing}} of their respective providers, as shown in Table~\ref{tab:cm}. In general, the input cost under cache hits is substantially lower than the standard input cost. Among the closed-source models considered, DeepSeek-Chat-V3.2 exhibits significantly lower input and output costs than the other models. Note that the cost weight $\lambda$ in the reward function is defined with respect to the per-token cost.
\section{Training and Inference Time and GPU Resources of DialRouter}
\begin{table}[H]
\caption{Time and GPU memory consumption for MCTS, training, and inference.}
\centering
\small
\begin{tabular}{lccc}
\toprule
 & MCTS & Training & Inference \\
\midrule
Time &
3h 40min (750 dialogues) &
1.3 s / batch &
0.03 s / decision \\
GPU Memory &
-- &
18.9 GB &
3636 MiB \\
\bottomrule
\end{tabular}
\label{tab:efficiency}
\end{table}
Training data are generated by running MCTS in parallel across dialogue instances. Processing 750 training dialogues for a single dataset takes approximately 3 hours and 40 minutes. All experiments are conducted on a single NVIDIA RTX~4090D GPU. DialRouter contains 472.9M parameters in total, with 44.42M trainable parameters. DialRouter is trained offline using the collected data. During training with a batch size of 32, the peak GPU memory usage is 18.9~GB, and the training throughput is approximately 1.3 seconds per batch. At inference time, routing decisions are made online, requiring 3636~MiB of GPU memory and only 0.03 seconds per decision (including a lightweight retrieval step taking approximately 0.01 seconds). Overall, DialRouter is lightweight and computationally efficient.

\section{Evaluator Preference Bias Analysis}
\begin{table}[h]
\caption{Task success rates evaluated by two independent evaluators (Qwen3-Max and GPT-5.1), with Spearman’s $\rho$ indicating consistent method rankings across domains.}
\centering
\small
\begin{tabular}{l|cccccc}
\toprule
& \multicolumn{2}{c}{ShareGPT ($\rho=0.9$)} & \multicolumn{2}{c}{JDDC ($\rho=0.7$)} & \multicolumn{2}{c}{MedDG ($\rho=0.9$)}\\
\cmidrule(lr){2-3} \cmidrule(lr){4-5} \cmidrule(lr){6-7}
\multirow{-2}{*}{Method} & Qwen3-Max & GPT-5.1 & Qwen3-Max & GPT-5.1 & Qwen3-Max & GPT-5.1\\ \midrule
Qwen3-14B&
83.24 & 77.77 & 77.19 & 78.76 & 72.50 & 70.17\\ 
KNN Router \cite{hu2024routerbench} &
83.37 & 76.45 & 78.44 & 78.67 & 71.99 & 68.74\\
Avengers \cite{zhang2025avengers} &
83.93 & 78.29 & 78.34 & 78.14 & 71.69 & 69.29\\ 
\textbf{DialRouter (Ours)} & 
87.46 & 81.01 & 83.45 & 84.39 & 79.03 & 76.26\\ 
MCTS Router & 
94.11 & 87.00 & 88.42 & 89.30 & 86.77 & 83.54\\ \bottomrule
\end{tabular}
\label{tab:g}
\end{table}
To assess robustness to evaluator choice and potential preference bias, we evaluate methods using two independent evaluators, Qwen3-Max and GPT-5.1. As shown in Table~\ref{tab:g}, method rankings remain largely consistent across all three dialogue domains, with high Spearman’s $\rho$ values. DialRouter consistently outperforms single LLMs and routing baselines under both evaluators, indicating that its performance gains are robust to evaluator-specific preferences.
\section{Analysis of DialRouter’s Routing Behavior}
\begin{figure}[H]
    \centering

    \begin{subfigure}{0.49\linewidth}
        \centering
        \includegraphics[width=\linewidth]{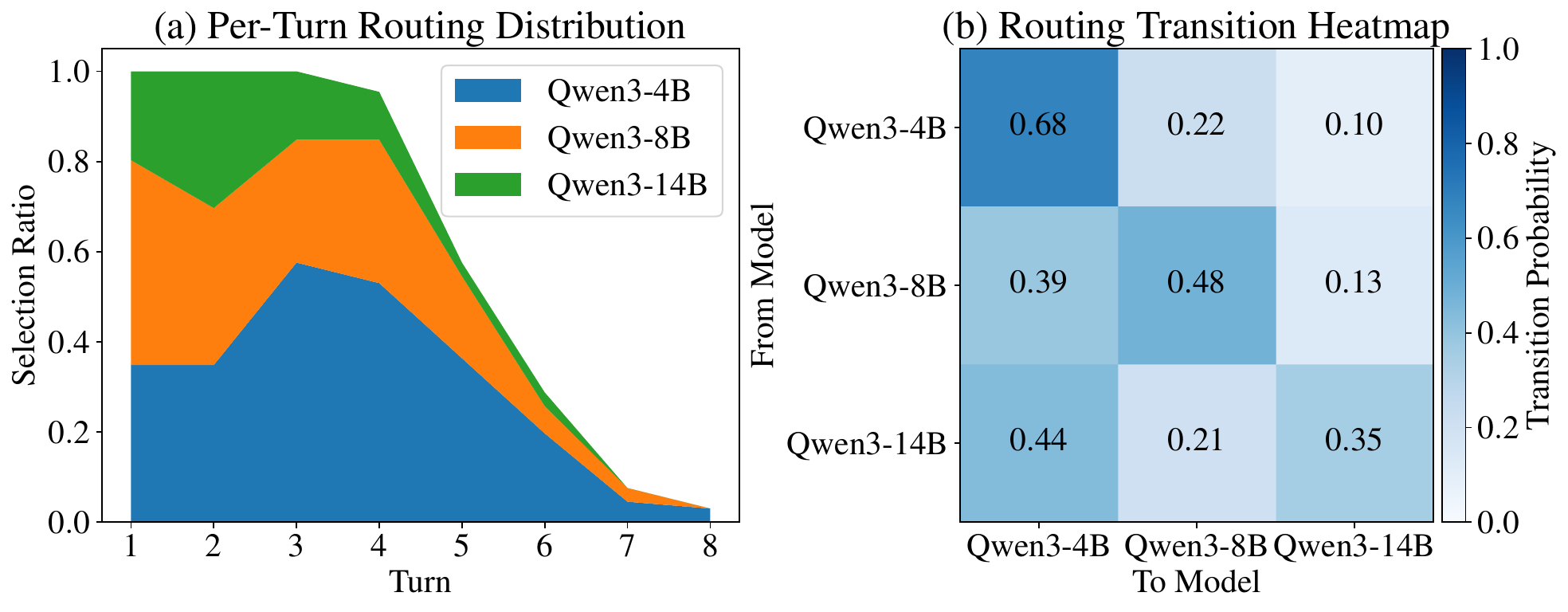}
        \caption{ShareGPT (Qwen)}
        \label{fig:a}
    \end{subfigure}\hfill
    \begin{subfigure}{0.49\linewidth}
        \centering
        \includegraphics[width=\linewidth]{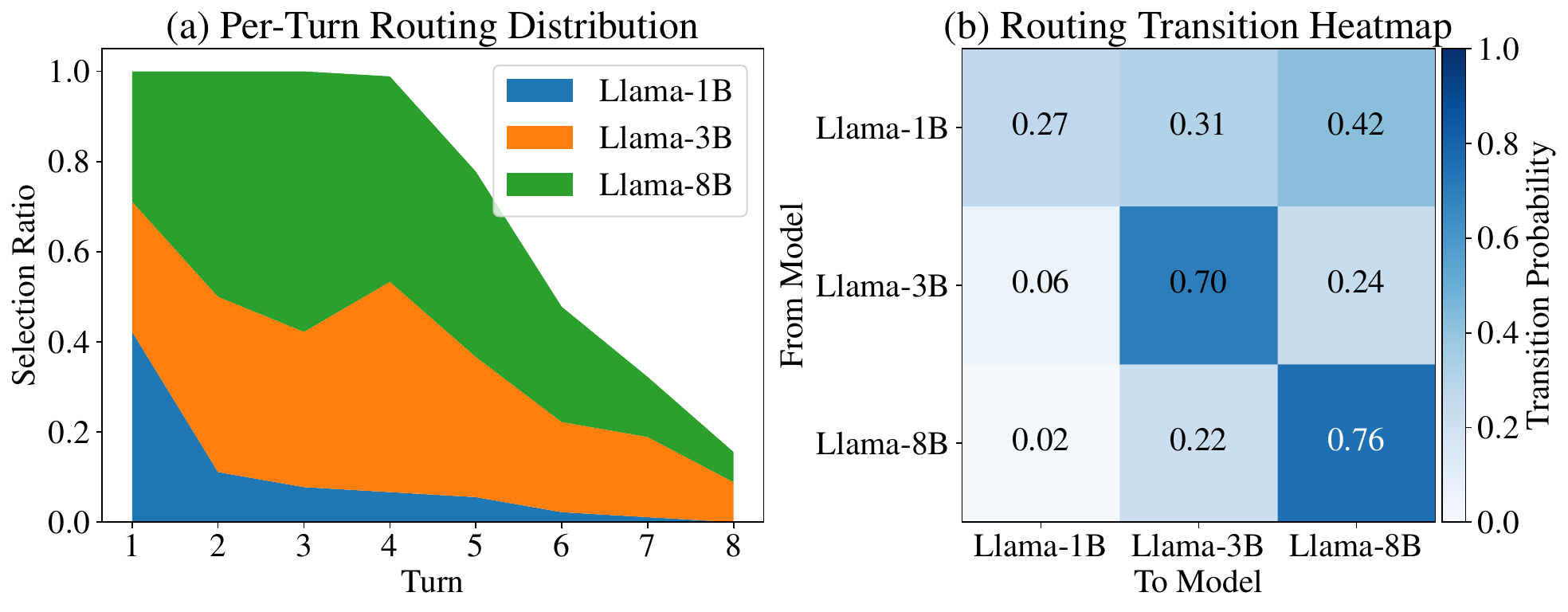}
        \caption{ShareGPT (Llama)}
        \label{fig:b}
    \end{subfigure}

    \vspace{0.5em}

    \begin{subfigure}{0.49\linewidth}
        \centering
        \includegraphics[width=\linewidth]{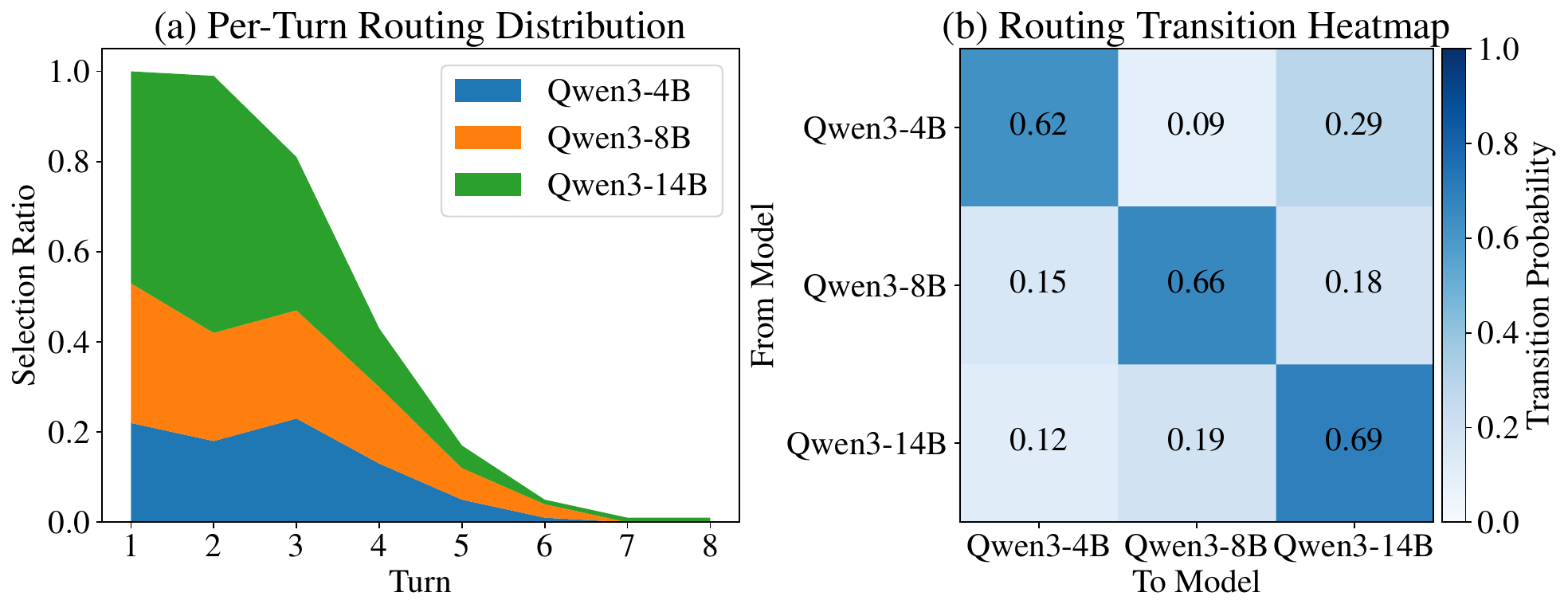}
        \caption{JDDC (Qwen)}
        \label{fig:c}
    \end{subfigure}\hfill
    \begin{subfigure}{0.49\linewidth}
        \centering
        \includegraphics[width=\linewidth]{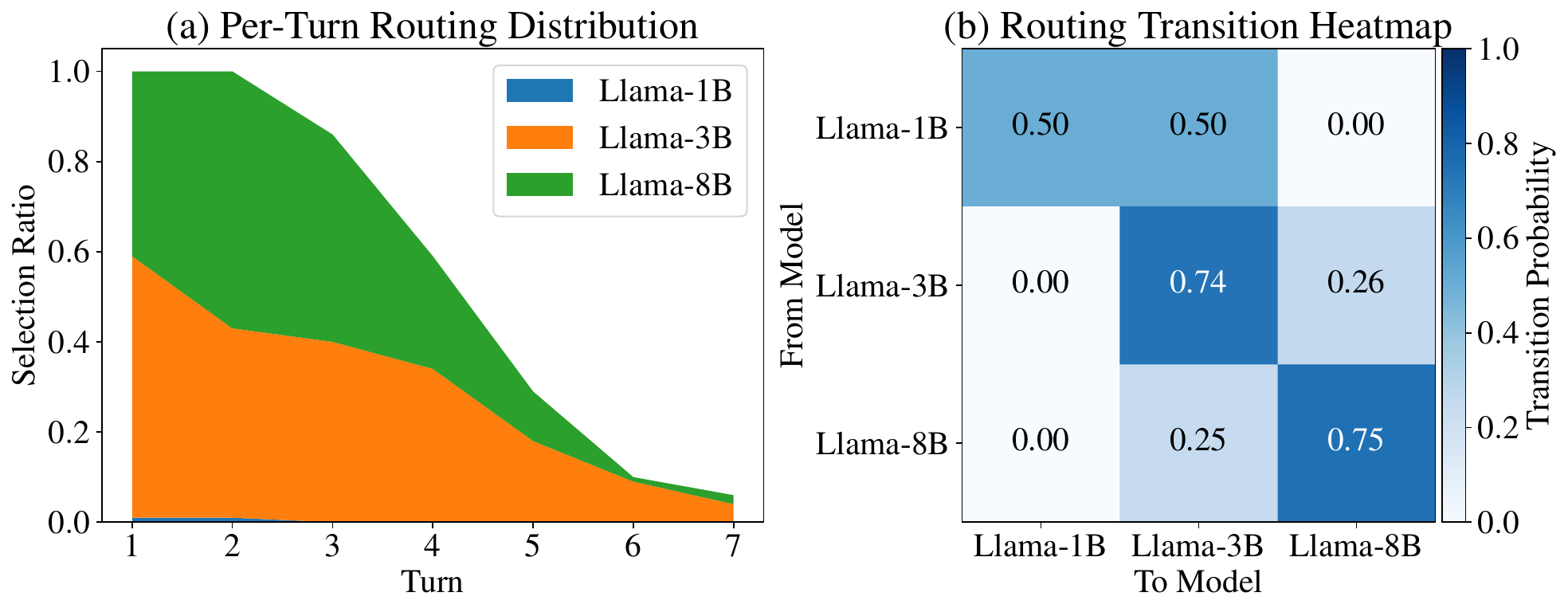}
        \caption{JDDC (Llama)}
        \label{fig:d}
    \end{subfigure}

    \vspace{0.5em}

    \begin{subfigure}{0.49\linewidth}
        \centering
        \includegraphics[width=\linewidth]{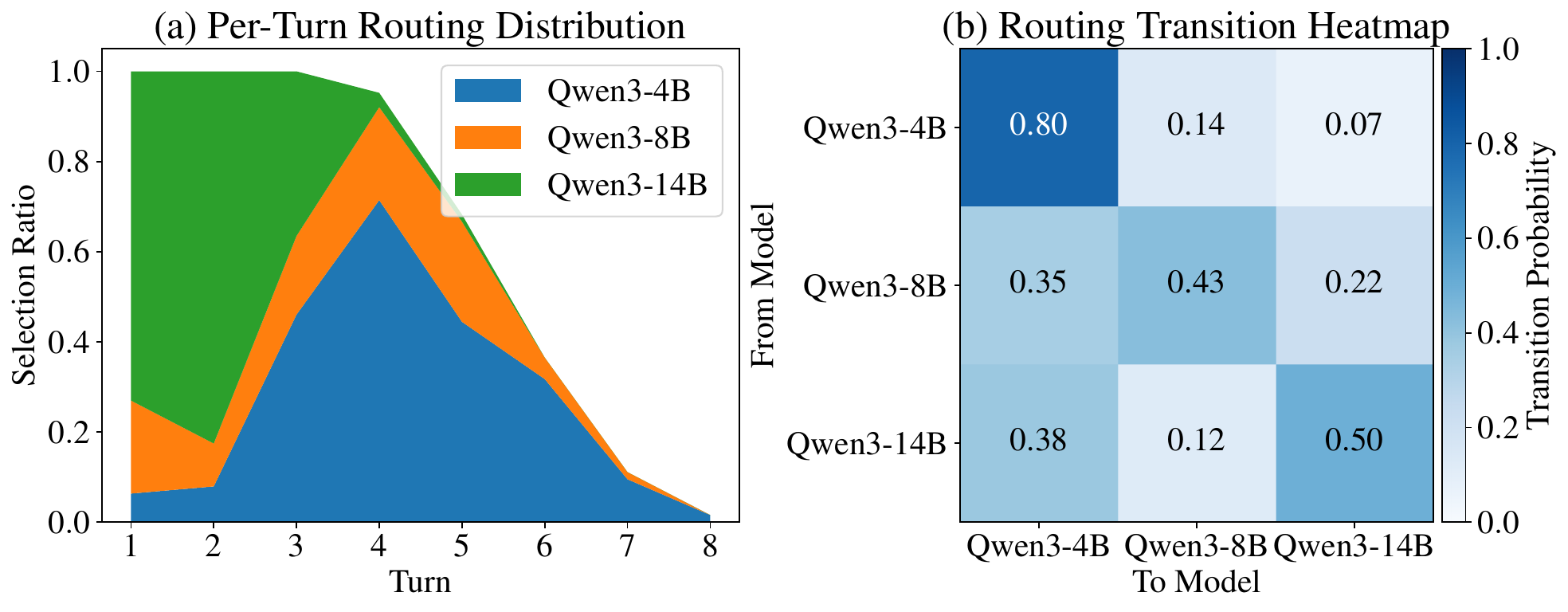}
        \caption{MedDG (Qwen)}
        \label{fig:e}
    \end{subfigure}\hfill
    \begin{subfigure}{0.49\linewidth}
        \centering
        \includegraphics[width=\linewidth]{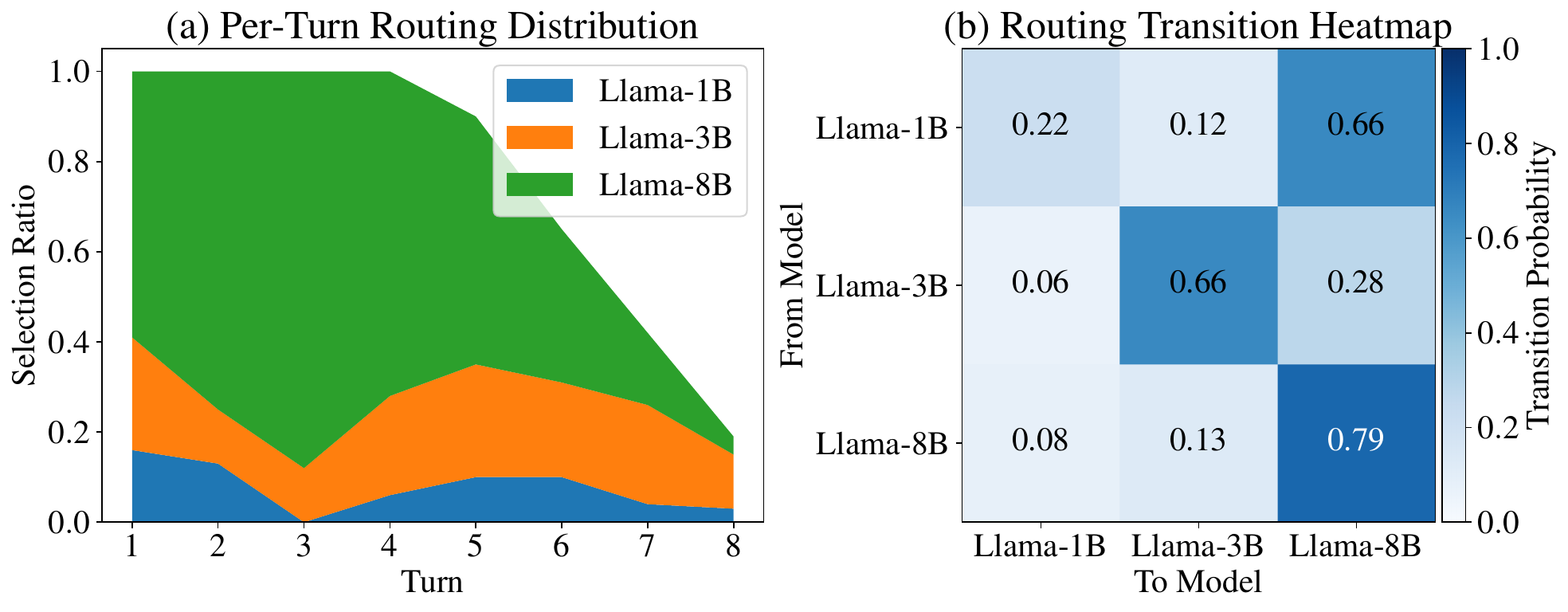}
        \caption{MedDG (Llama)}
        \label{fig:f}
    \end{subfigure}

    \caption{Per-turn routing selection distributions (stacked proportion plots) and inter-turn routing transition heatmaps for the Qwen (left) and Llama (right) candidate sets across three datasets. The stacked areas show the proportion of dialogue trajectories selecting each model at each turn; the total height decreases in later turns as fewer dialogues extend to longer horizons.}
    \label{fig:6imgs}
\end{figure}

This section analyzes routing behaviors across different dialogue scenarios by visualizing multi-turn LLM routing trajectories. As shown in Figure\ref{fig:6imgs}, we present per-turn LLM selection distributions (stacked proportion plots) and inter-turn transition patterns (transition matrices) on three datasets to characterize the dynamic evolution of routing strategies over dialogue stages. By comparing different candidate LLM sets within the same dataset, we examine how routing adapts to task contexts and dialogue phases, highlighting differences in exploration, convergence, and stability.

On \textbf{ShareGPT}, the Qwen candidate set favors medium-sized models in early turns and gradually shifts toward smaller models as the dialogue progresses, with the transition matrix indicating frequent fallback transitions from larger to smaller models, resulting in a clear strong-to-weak convergence pattern. In contrast, the Llama candidate set more often maintains or returns to larger models in later turns, with higher self-loop and back-transition probabilities, suggesting sustained reliance on high-capacity models in open-domain multi-turn conversations.

On \textbf{JDDC}, the Qwen candidate set quickly converges to large models in the initial turns and remains relatively stable thereafter, with a transition structure dominated by self-loops, indicating limited cross-model switching once a routing strategy is established. By comparison, the Llama candidate set is largely dominated by medium and large models in early turns and exhibits noticeable transitions from medium to large models, reflecting a preference for high-capacity models in task-constrained e-commerce dialogue scenarios.

On \textbf{MedDG}, the Qwen candidate set strongly prefers large models at the beginning of the dialogue, followed by a stage-wise shift toward medium and smaller models in later turns, with frequent transitions from large to medium models, capturing strategy adjustments during information gathering and verification phases in medical conversations. In contrast, the Llama candidate set consistently concentrates on large models across most turns, with dominant self-loop transitions, indicating sustained reliance on high-capacity models under high-stakes and safety-critical medical settings.

\section{Prompting Details}\label{appenb}
We use Qwen3-Max as both the user simulator and the reward model, and design corresponding system prompts for ShareGPT, JDDC, and MedDG, which are instantiated with the user profile $p_{\text{usr}}$ and the checklist $C$ extracted from each dialogue instance. The user simulator and evaluator of open-domain ShareGPT are not tied to a specific domain, enabling a general and reusable simulation and evaluation for new domains. The detailed system prompts are illustrated in Figures~\ref{table:sgu}--\ref{table:dge}.
\begin{figure}[H]
\centering
\begin{tcolorbox}[title=System Prompt for the ShareGPT User Simulator]
\footnotesize 
You are a human user engaging in a multi-turn conversation with an AI assistant. Please generate user utterances that are natural, coherent, and aligned with the specified conversational goals.\\

\#\#\# Conversation Goals

- User intents: [User Intents]\\

Please follow the rules below throughout the conversation:

1. In each turn, your utterance should be guided by the currently triggered intent. Do not introduce goals that are unrelated to the specified intents or deviate from the conversation scope.

2. Your response should naturally follow the assistant’s previous reply and maintain conversational coherence. You may exhibit realistic user behaviors such as asking follow-up questions, seeking clarification, or restating information when appropriate.

3. Use natural, conversational language rather than structured or list-style expressions, and keep each utterance at a moderate length.

4. If all intents have been fully satisfied, or no further intent can be reasonably triggered, terminate the conversation by outputting: \textless End of Conversation\textgreater.
\end{tcolorbox}
\caption{System prompt for the ShareGPT user simulator.}\label{table:sgu}
\end{figure}

\begin{figure}[H]
\centering
\begin{tcolorbox}[colback=orange!5!white, colframe=orange!50!black, title=System Prompt for the ShareGPT Reward Model]
\footnotesize 
You are a professional dialogue evaluator. Your task is to assess the assistant’s performance in the conversation based on a given checklist and determine whether the dialogue has terminated.\\

\#\#\# Dialogue History

[Dialogue History]\\

\#\#\# Checklist

[Checklist]\\

\#\#\# Scoring Guidelines

For each checklist item, evaluate whether the assistant has completed the corresponding task behavior in the dialogue and assign a score according to the following criteria:

- 1: The assistant clearly and fully satisfies the item, providing complete and explicit information aligned with the user’s intent;

- 0.5: The assistant partially satisfies the item, but the response is vague, incomplete, or only indirectly addresses the requirement;

- 0: The assistant does not address the item at all or clearly fails to fulfill the requirement.

All judgments must be strictly based on the observed dialogue content and the behaviors specified in the checklist. Do not speculate, infer, or rationalize any assistant behavior that is not explicitly present in the dialogue.\\

\#\#\# Dialogue Termination Determination

The dialogue is considered terminated, and the field `done` should be set to 1, if and only if the user’s final complete utterance exactly matches the string “\textless End of Conversation\textgreater”. Otherwise, set `done` to 0.\\

\#\#\# Output Format

Return only the following JSON object, without any additional text or explanation:

\{

  "checklist": [score\_1, score\_2, ..., score\_n],

  "done": 0 or 1

\}
\end{tcolorbox}
\caption{System prompt for the ShareGPT reward model.}\label{table:sge}
\end{figure}

\begin{figure}[H]
\centering
\begin{tcolorbox}[title=System Prompt for the JDDC User Simulator (English)]
\footnotesize
You are an e-commerce platform user. Based on the given user background and goals, please engage in a multi-turn conversation with the customer-service system.\\

\#\#\# User Behavioral Patterns

During the conversation, please consistently exhibit the following behavioral patterns:

- Behavioral patterns: [Behavior Patterns]\\

\#\#\# User General Context

- Background information: [General Context]\\

\#\#\# User Goals

- Dialogue goals: [User Goals]\\

Please strictly follow the dialogue rules below:\\

1. You may only act as the user. Do not simulate, replace, or guide the customer-service agent’s responses.

2. In each turn, generate a natural user utterance based on the customer-service agent’s previous reply, using language and tone consistent with real e-commerce interactions.

3. Your utterances should remain focused on the specified user goals. Do not introduce unrelated topics or requests.

4. Once you believe that sufficient information has been obtained to fully satisfy the user goals, you may directly terminate the conversation without waiting for additional confirmation or actions.

5. If the customer-service agent explicitly indicates that the request cannot be further processed, or you determine that all goals have been completed, output: \textless End of Conversation\textgreater.
\end{tcolorbox}

\begin{tcolorbox}[title=System Prompt for the JDDC User Simulator (Chinese)]
\footnotesize 
\begin{CJK}{UTF8}{gbsn}
你是一名电商平台用户，请根据给定的用户背景和目标，与客服系统展开多轮对话。\\

\#\#\# 用户行为特征

在对话过程中，请始终体现以下行为倾向：

- 行为特征：[用户行为特征]\\

\#\#\# 用户背景

- 背景信息：[用户通用背景]\\

\#\#\# 用户目标

- 对话目标：[用户目标]\\

请你严格遵循以下对话规则：\\

1. 你只能扮演用户角色，不要模拟、代替或引导客服的发言。

2. 在每一轮对话中，请基于客服上一轮的回复内容，自然地生成当前轮的用户发言，语气应符合真实电商用户的交流习惯。

3. 你的发言应始终围绕用户目标展开，不要主动引入与目标无关的话题或需求。

4. 当你认为已获取到满足目标的完整信息时，可以直接结束对话，无需等待额外确认或操作。

5. 如果客服明确表示无法继续处理你的请求，或你判断所有目标均已完成，请输出：\textless结束对话\textgreater。
\end{CJK}
\end{tcolorbox}
\caption{System prompt for the JDDC user simulator.}\label{table:jdu}
\end{figure}

\begin{figure}[H]
\centering
\begin{tcolorbox}[colback=orange!5!white, colframe=orange!50!black, title=System Prompt for the JDDC Reward Model (English)]
\footnotesize
You are a professional customer-service dialogue evaluator. Your task is to assess the customer-service agent’s behavior throughout the conversation based on the given checklist and to determine whether the dialogue has ended.\\

\#\#\# Dialogue History

[Dialogue History]\\

\#\#\# Checklist

[Checklist]\\

\#\#\# Scoring Guidelines

For each item in the checklist, evaluate the customer-service agent’s behavior in the dialogue and assign a score according to the following guidelines:

- 1: The agent clearly and fully completed the corresponding service behavior;

- 0.5: The agent partially completed the service behavior, but the response is unclear, insufficient, or only indirectly addresses the requirement;

- 0: The agent did not address the service behavior at all or clearly failed to fulfill the requirement.

When assigning scores, strictly base your judgment on the content explicitly present in the dialogue. Do not rely on common sense, assumptions, or speculative reasoning to fill in missing information.\\

\#\#\# Dialogue Termination Determination

The dialogue is considered terminated if and only if the user’s final complete utterance exactly matches the string “\textless End of Conversation\textgreater”. In this case, set `done` to 1; otherwise, set `done` to 0.\\

\#\#\# Output Format

Return the evaluation result strictly in the following JSON format, without any additional text or explanation:

\{

  "checklist": [score\_1, score\_2, ...],

  "done": 0 or 1

\}
\end{tcolorbox}

\begin{tcolorbox}[colback=orange!5!white, colframe=orange!50!black, title=System Prompt for the JDDC Reward Model (Chinese)]
\footnotesize 
\begin{CJK}{UTF8}{gbsn}
你是一名专业的客服对话评价器。你的任务是依据给定的 Checklist，对客服在整个对话过程中的行为进行逐项评估，并判断该对话是否已经结束。\\

\#\#\# 对话历史

[对话历史]\\

\#\#\# Checklist

[Checklist]\\

\#\#\# 评分标准

请针对 Checklist 中的每一项内容，结合对话历史进行判断，并给出对应评分：

- 1：客服明确且充分地完成了该项服务行为；

- 0.5：客服部分完成了该项服务行为，但表达不清晰、不充分，或仅以间接方式涉及；

- 0：客服完全未涉及该项服务行为，或明显未完成该项要求。

评分时请严格依据对话中实际出现的内容进行判断，不得基于常识或主观推断对缺失信息进行合理化补全。\\

\#\#\# 对话终止判定

若且仅若用户在最后一轮对话中的完整回复严格等于字符串“\textless结束对话\textgreater”，则认为该对话已经结束，并将 done 设为 1；否则，将 done 设为 0。\\

\#\#\# 输出格式

请仅按照以下 JSON 格式输出评价结果，不得包含任何额外文本或解释说明：

\{

  "checklist": [评分1, 评分2, …],
  
  "done": 0 或 1
  
\}
\end{CJK}
\end{tcolorbox}
\caption{System prompt for the JDDC reward model.}\label{table:jde}
\end{figure}

\begin{figure}[H]
\centering
\begin{tcolorbox}[title=System Prompt for the MedDG User Simulator (English)]
\footnotesize
You are a patient participating in an online medical consultation. Based on the given patient background and consultation goals, please engage in a multi-turn conversation with the medical system.\\

\#\#\# Patient Information

- Age: [Age]

- Gender: [Gender]

- Chief complaint: [Chief Complaint]

- Trigger or onset: [Suspected Trigger]

- Medical history: [Medical History]

- Current medication: [Current Medication]\\

\#\#\# Symptom Details

- Detailed description of current symptoms: {Symptom Details}\\

\#\#\# Consultation Goals

- Consultation goals: [User Goals]\\

Please strictly follow the dialogue rules below:\\

1. You may only act as the patient. Do not simulate or replace the doctor’s role.

2. In each turn, generate a natural, realistic, and logically coherent patient utterance based on the doctor’s previous response.

3. You should exhibit a help-seeking and awaiting-professional-judgment patient stance. Do not proactively correct, confirm, or explain the doctor’s medical conclusions.

4. Your utterances should remain focused on the consultation goals. Do not introduce topics unrelated to the goals or deviate from the current consultation theme.

5. You may terminate the conversation only when you are confident that the consultation goals have been sufficiently satisfied, or when the doctor explicitly indicates that no further assistance can be provided. Before deciding to end the conversation, please ensure that: The consultation goals have been clearly and accurately addressed by the doctor; The doctor’s advice is explicit and actionable; No remaining questions or ambiguities require further clarification.

6. When you determine that the conversation has met the termination conditions, output: \textless End of Conversation\textgreater.
\end{tcolorbox}

\begin{tcolorbox}[title=System Prompt for the MedDG User Simulator (Chinese)]
\footnotesize 
\begin{CJK}{UTF8}{gbsn}
你是一名正在进行在线问诊的患者，请根据给定的病人背景和就诊目标，与医生系统展开多轮对话。\\

\#\#\# 患者基本信息

- 年龄：[年龄]

- 性别：[性别]

- 主诉：[主要不适或症状]

- 诱因或起病情况：[可能的诱发因素]

- 既往病史：[既往相关病史]

- 当前用药情况：[是否正在用药及用药情况]\\

\#\#\# 症状细节

- 当前症状的具体描述：[症状细节描述]\\

\#\#\# 就诊目标

- 就诊目标：[希望通过本次问诊获得的信息或建议]\\

请你严格遵循以下对话规则：\\

1. 你只能扮演患者，不得模拟或代替医生角色。

2. 在每一轮对话中，请基于医生上一轮的回复内容，生成自然、真实且具有逻辑一致性的患者发言。

3. 你应表现出寻求帮助、等待专业判断的患者状态，不要主动纠正、确认或解释医生的专业结论。

4. 你的发言应始终围绕就诊目标展开，不要主动引入与目标无关的问题，也不要偏离当前就诊主题。

5. 只有在你确认就诊目标已被充分满足，或医生明确表示无法继续提供帮助时，才可以结束对话。在判断是否结束前，请确认：就诊目标是否已被医生清晰、准确地回应；医生给出的建议是否明确且具有可操作性；是否仍存在需要进一步澄清的问题。

6. 当你确认对话已满足结束条件时，你的发言应为：\textless结束对话\textgreater。
\end{CJK}
\end{tcolorbox}
\caption{System prompt for the MedDG user simulator.}\label{table:dgu}
\end{figure}

\begin{figure}[H]
\centering
\begin{tcolorbox}[colback=orange!5!white, colframe=orange!50!black, title=System Prompt for the MedDG Reward Model (English)]
\footnotesize
You are a professional medical dialogue evaluator. Your task is to assess the doctor’s performance throughout the medical consultation based on the given dialogue history and checklist, and to determine whether the consultation has ended.\\

\#\#\# Dialogue History

[Dialogue History]\\

\#\#\# Checklist

[Checklist]\\

\#\#\# Scoring Guidelines

Based on the dialogue history, evaluate whether the doctor has completed each item in the checklist and assign a score according to the following criteria:

- 1: The doctor has clearly and fully completed the item, with information that is explicit, complete, and medically sound;

- 0.5: The doctor has partially addressed the item, but the response is unclear, insufficient, or only indirectly related;

- 0: The doctor did not address the item at all or clearly failed to fulfill the corresponding consultation responsibility.

When assigning scores, strictly base your judgment on the content explicitly present in the dialogue and the behaviors required by the checklist. Do not make subjective assumptions or rationalize missing information that is not explicitly provided by the doctor.\\

\#\#\# Dialogue Termination Determination

The consultation is considered terminated if and only if the patient’s final complete utterance exactly matches the string “<End of Conversation>”. In this case, set `done` to 1; otherwise, set `done` to 0.\\

\#\#\# Output Format

Return the evaluation result strictly in the following JSON format, without any additional text or explanation:

\{

  "checklist": [score\_1, score\_2, ...],

  "done": 0 or 1

\}
\end{tcolorbox}

\begin{tcolorbox}[colback=orange!5!white, colframe=orange!50!black, title=System Prompt for the MedDG Reward Model (Chinese)]
\footnotesize 
\begin{CJK}{UTF8}{gbsn}
你是一名专业的医疗问诊对话评价器。你的任务是基于给定的对话历史和 Checklist，逐项评估医生在问诊过程中的表现，并判断问诊是否已经结束。\\

\#\#\# 对话历史

[对话历史]\\

\#\#\# Checklist

[Checklist]\\

\#\#\# 评分标准

请根据对话历史，逐项判断医生是否完成 Checklist 中的每一项内容，并按照以下标准进行评分：

- 1：医生已明确且充分完成该项内容，信息表达清晰、完整，且在医学上合理；

- 0.5：医生对该项内容有所涉及，但表达不够清晰、信息不充分，或仅作了间接回应；

- 0：医生完全未涉及该项内容，或明显未履行相应的问诊职责。

评分时请严格依据实际对话内容与 Checklist 所要求的服务行为进行判断，不得进行主观推测，也不得合理化补全医生未明确给出的信息。\\

\#\#\# 对话终止判定

若且仅若患者在最后一轮对话中的完整回复严格等于字符串“\textless结束对话\textgreater”，则认为该对话已经结束，并将 done 设为 1；否则，将 done 设为 0。\\

\#\#\# 输出格式

请仅按照以下 JSON 格式输出评价结果，不得包含任何额外文本或解释说明：

\{

  "checklist": [评分1, 评分2, …],
  
  "done": 0 或 1
  
\}
\end{CJK}
\end{tcolorbox}
\caption{System prompt for the MedDG reward model.}\label{table:dge}
\end{figure}

\section{Case Studies}
\begin{figure}[H]
    \centering
    \includegraphics[width=1.0\linewidth]{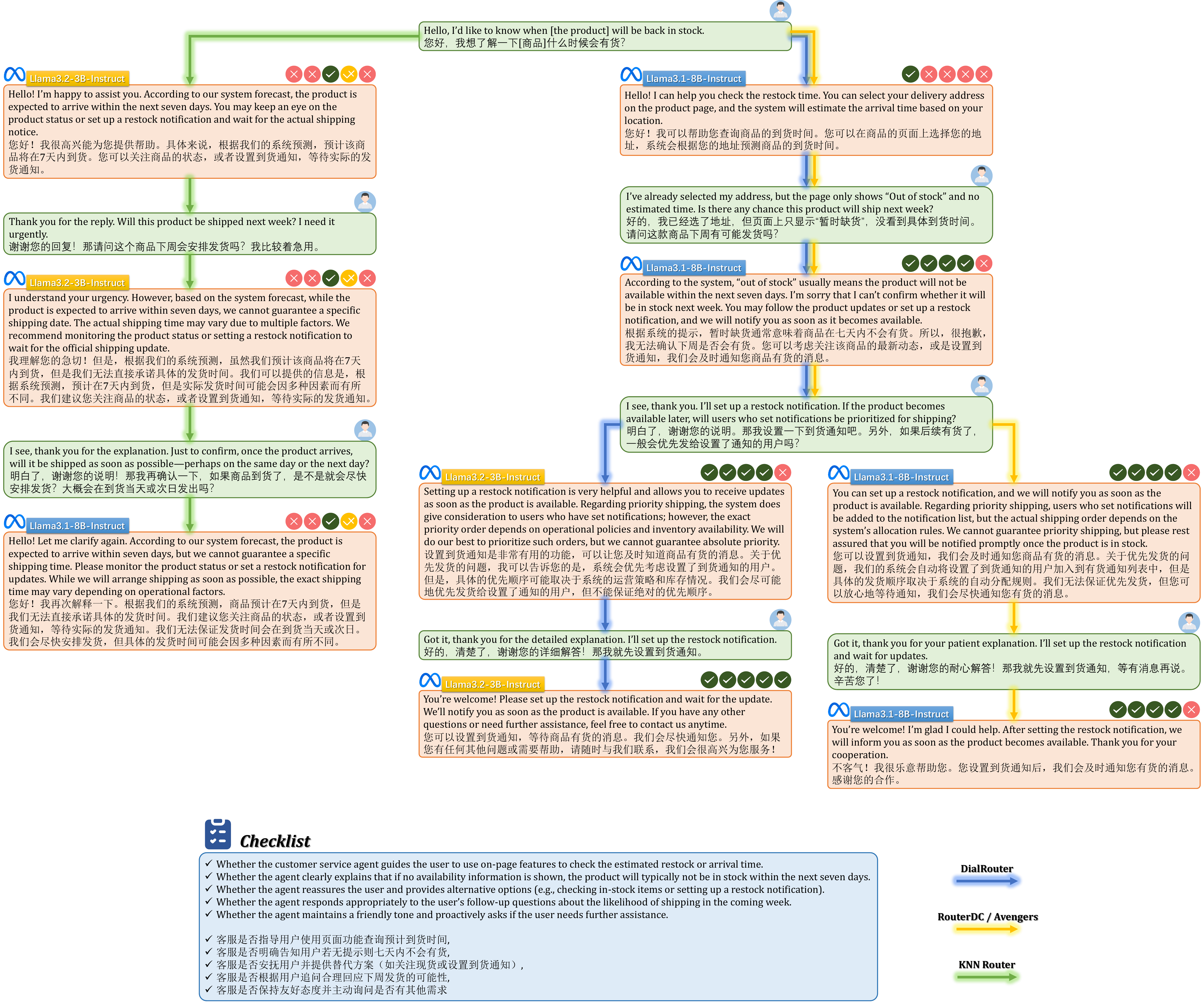}
    \caption{Case study on multi-turn e-commerce customer service in JDDC with Llama candidate set.}
    \label{fig:c2}
\end{figure}

\begin{figure}[H]
    \centering
    \includegraphics[width=1.0\linewidth]{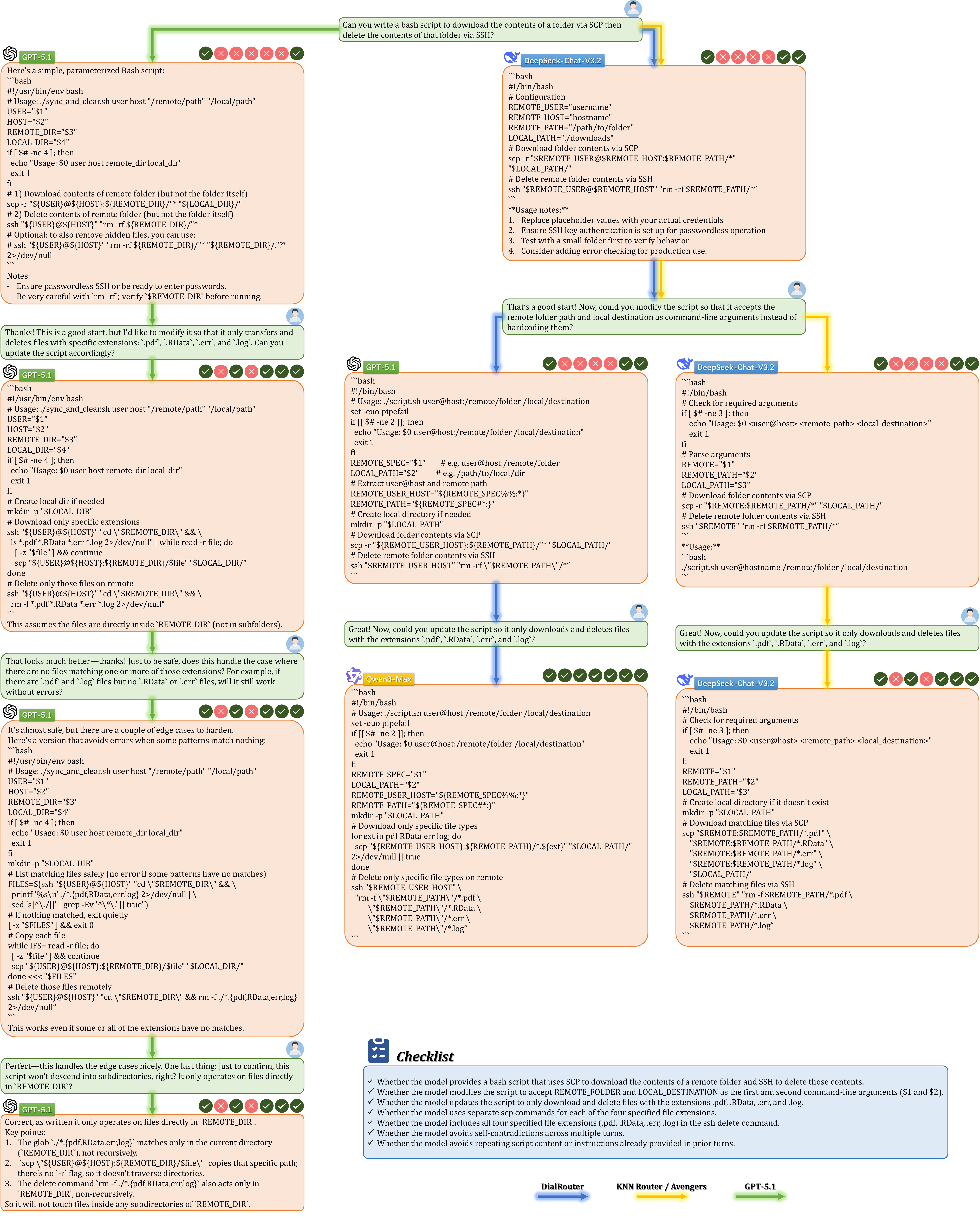}
    \caption{Case study on multi-turn script writing in ShareGPT with closed-source LLM candidate set.}
    \label{fig:c1}
\end{figure}

\begin{figure}[H]
    \centering
    \includegraphics[width=1.0\linewidth]{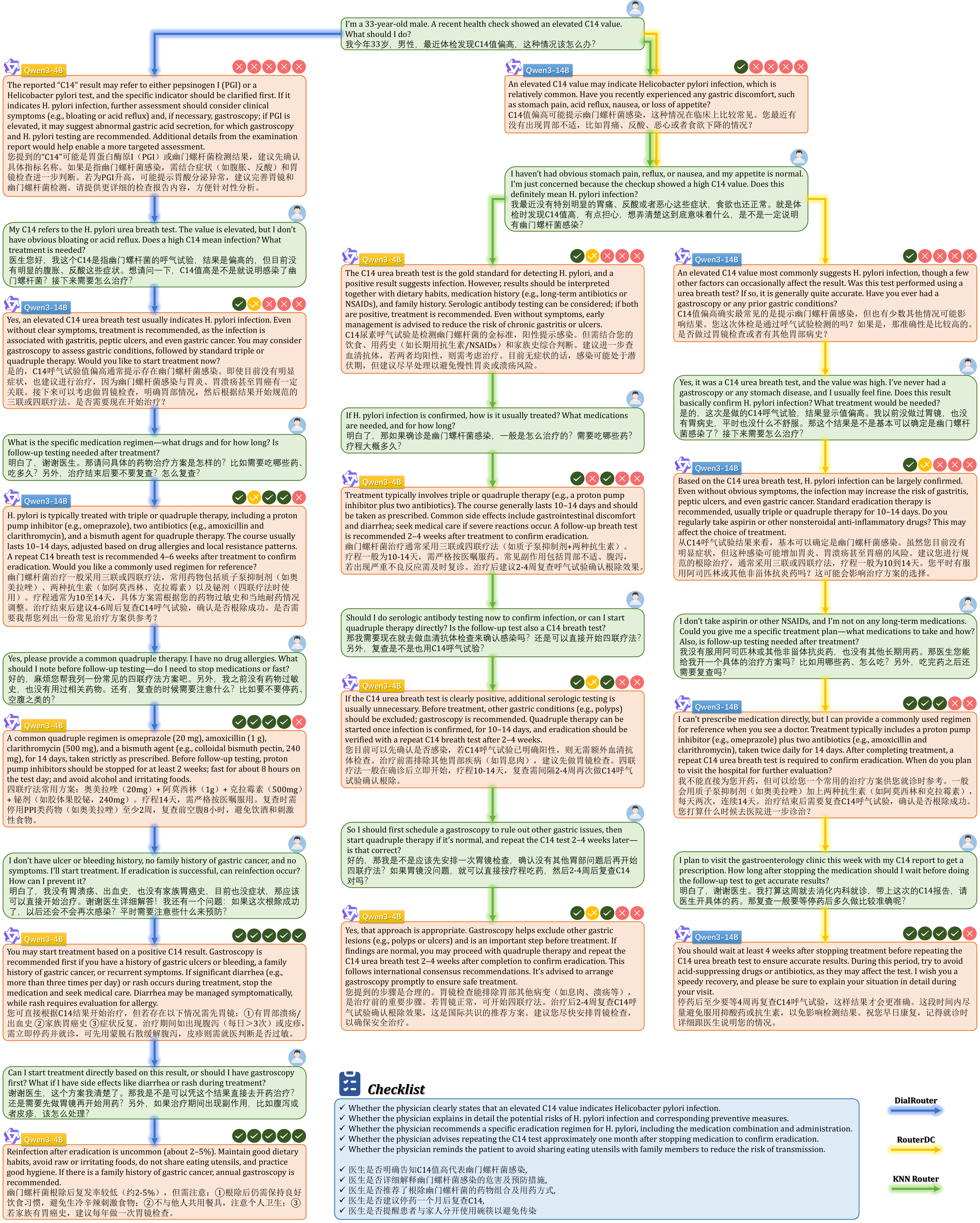}
    \caption{Case study on multi-turn medical consultation in MedDG with Qwen candidate set.}
    \label{fig:c3}
\end{figure}

As shown in Figures~\ref{fig:c1},~\ref{fig:c3}, and~\ref{fig:c2}, the three case studies cover distinct multi-turn dialogue scenarios, including script writing in ShareGPT, medical consultation in MedDG, and e-commerce customer service in JDDC, where user requirements progressively evolve and become more specific as the dialogue unfolds. Across these scenarios, DialRouter is able to perceive the dialogue stage and select models that better match the current requirement structure, thereby maintaining decision consistency across turns and incrementally completing key information. By employing higher-capacity models for high-risk or critical decision stages and more lightweight or stable models for execution details and supplementary explanations, DialRouter ultimately achieves full coverage of the checklist items. In contrast, some baseline methods rely on a fixed model or lack stage awareness, leading them to overlook newly introduced constraints or details in later turns and fail to fully satisfy dialogue-level objectives, while others switch models without a holistic understanding of dialogue stages and risk, resulting in oscillations or drift along critical decision paths.

Overall, these cases consistently demonstrate that success in multi-turn dialogue does not depend on producing the strongest possible response at each individual turn, but rather on the router’s ability to perceive dialogue stages and make long-horizon model selection decisions, thereby jointly ensuring directional correctness, detail completeness, and risk control throughout the evolving interaction and ultimately achieving stable dialogue-level goal fulfillment.


\end{document}